\documentclass[letterpaper]{article}
\usepackage{proceed}
\usepackage[margin=1in]{geometry}

\usepackage{chicago}
\usepackage{times}
\usepackage{xcolor} 
\usepackage{graphicx}
\usepackage{hyperref}
\usepackage{listings}
\usepackage{amsmath}
\usepackage{amsfonts}

\newtheorem{THEOREM}{Theorem}[section]
\newenvironment{theorem}{\begin{THEOREM} \hspace{-.85em} {\bf :} }%
                        {\end{THEOREM}}
\newtheorem{LEMMA}[THEOREM]{Lemma}
\newenvironment{lemma}{\begin{LEMMA} \hspace{-.85em} {\bf :} }%
                      {\end{LEMMA}}
\newtheorem{COROLLARY}[THEOREM]{Corollary}
\newenvironment{corollary}{\begin{COROLLARY} \hspace{-.85em} {\bf :} }%
                          {\end{COROLLARY}}
\newtheorem{PROPOSITION}[THEOREM]{Proposition}
\newenvironment{proposition}{\begin{PROPOSITION} \hspace{-.85em} {\bf :} }%
                            {\end{PROPOSITION}}
\newtheorem{DEFINITION}[THEOREM]{Definition}
\newenvironment{definition}{\begin{DEFINITION} \hspace{-.85em} {\bf :} \rm}%
                            {\end{DEFINITION}}
\newtheorem{CLAIM}[THEOREM]{Claim}
\newenvironment{claim}{\begin{CLAIM} \hspace{-.85em} {\bf :} \rm}%
                            {\end{CLAIM}}
\newtheorem{EXAMPLE}[THEOREM]{Example}
\newenvironment{example}{\begin{EXAMPLE} \hspace{-.85em} {\bf :} \rm}%
                            {\end{EXAMPLE}}
\newtheorem{REMARK}[THEOREM]{Remark}
\newenvironment{remark}{\begin{REMARK} \hspace{-.85em} {\bf :} \rm}%
                            {\end{REMARK}}

\newcommand{\thm}{\begin{theorem}}
\newcommand{\lem}{\begin{lemma}}
\newcommand{\pro}{\begin{proposition}}
\newcommand{\dfn}{\begin{definition}}
\newcommand{\rem}{\begin{remark}}
\newcommand{\xam}{\begin{example}}
\newcommand{\cor}{\begin{corollary}}
\newcommand{\prf}{\noindent{\bf Proof:} }
\newcommand{\ethm}{\end{theorem}}
\newcommand{\elem}{\end{lemma}}
\newcommand{\epro}{\end{proposition}}
\newcommand{\edfn}{\bbox\end{definition}}
\newcommand{\erem}{\bbox\end{remark}}
\newcommand{\exam}{\bbox\end{example}}
\newcommand{\ecor}{\end{corollary}}
\newcommand{\eprf}{\bbox\vspace{0.1in}}
\newcommand{\beqn}{\begin{equation}}
\newcommand{\eeqn}{\end{equation}}

\newcommand{\bbox}{\vrule height7pt width4pt depth1pt}

\newcommand{\clm}{\begin{claim}}
\newcommand{\eclm}{\end{claim}}

\newcommand{\sat}{\models}

\newcommand{\union}{\cup}

\newcommand{\IR}{\mbox{$I\!\!R$}}

\renewcommand{\phi}{\varphi}

\newcommand{\A}{{\cal A}}
\newcommand{\B}{{\cal B}}

\newcommand{\F}{{\cal F}}

\newcommand{\I}{{\cal I}}

\newcommand{\M}{{\cal M}}

\newcommand{\R}{{\cal R}}

\newcommand{\U}{{\cal U}}
\newcommand{\V}{{\cal V}}

\newcommand{\ol}{\setlength{\itemsep}{0pt}\begin{enumerate}}
\newcommand{\eol}{\end{enumerate}\setlength{\itemsep}{-\parsep}}
\newcommand{\ul}{\setlength{\itemsep}{0pt}\begin{itemize}}
\newcommand{\dl}{\setlength{\itemsep}{0pt}\begin{description}}
\newcommand{\edl}{\end{description}\setlength{\itemsep}{-\parsep}}
\newcommand{\eul}{\end{itemize}\setlength{\itemsep}{-\parsep}}

\newcommand{\commentout}[1]{}

\newcommand{\bi}{\begin{itemize}}
\newcommand{\ei}{\end{itemize}}
\newcommand{\be}{\begin{enumerate}}
\newcommand{\ee}{\end{enumerate}}

\renewcommand{\S}{{\cal S}}
\newcommand{\Rest}{\mathit{Rst}}

\newcommand{\fullv}[1]{#1}
\newcommand{\shortv}{\commentout}

\begin{document}

\title{Approximate Causal Abstraction}

\author{Sander Beckers\\
  Dept. of Philosophy and Religious Studies\\
  Utrecht University\\
  srekcebrednas@gmail.com
  \And
  Frederick Eberhardt\\
  Humanities and Social Sciences\\
  California Instituite of Technology\\
  fde@caltech.edu
  \And
  Joseph Y. Halpern\\
  Dept. of Computer Science\\
  Cornell University\\
  halpern@cs.cornell.edu}


\maketitle

\begin{abstract}
  Scientific models describe natural phenomena at different levels of
  abstraction. Abstract descriptions can provide the basis for
  interventions on the system and explanation of observed phenomena at a
  level of granularity that is coarser than the most fundamental account
  of the system. Beckers and Halpern~\citeyear{BH19}, building on 
  work of
  Rubenstein et al.~\citeyear{Rub17}, developed an account of
  \emph{abstraction} for causal models that is exact. Here we extend
  this account to the more realistic case where an abstract causal model
  offers only an approximation of the underlying system. We show how the
  resulting account handles the discrepancy that can arise between low-
  and high-level causal models of the same system, and in the process provide
  an account of how one causal model approximates another, a topic of
  independent interest. Finally, we extend the account of approximate
  abstractions to probabilistic causal models, indicating how and where
  uncertainty can enter into an approximate abstraction. 

\end{abstract}

\section{INTRODUCTION}

Scientific models aim to provide a description of reality that offers
both an explanation of observed phenomena and
a basis for intervening on and manipulating the system to bring
about desired outcomes. Both of these aims lead to a consideration of
models that represent the \emph{causal} relations governing the
system.  They 
also imply the need for scientific models that describe the system at
a granularity or level of description appropriate for the user and suitable for interventions that are feasible. Such more
\emph{abstract} causal models do not capture all the detailed
interactions that occur at the most fundamental level of the system, nor do they, in general,
represent outcomes of the system completely accurately at the abstract
level. Nevertheless, such abstract causal models can (at least)
\emph{approximately} explain the phenomena, and can be informative
about how the system will respond to interventions that are specified
only at the abstract level. 

This paper provides a formal account of such approximate abstractions for
causal models that builds on the definition of an 
\emph{abstraction} 
provided by Beckers and Halpern \citeyear{BH19} (see Section~\ref{sec:review}), 
which in turn built on the work of Rubenstein et al.~\citeyear{Rub17}.
That notion of abstraction implicitly assumed an underlying causal
system that permitted an exact description of the system at the abstract level. 
Here we weaken that 
assumption to handle what we take to be the more realistic case,
namely, that abstract causal models will  capture the underlying
system in only an approximate way.

\begin{figure}
  \includegraphics[width = 8cm]{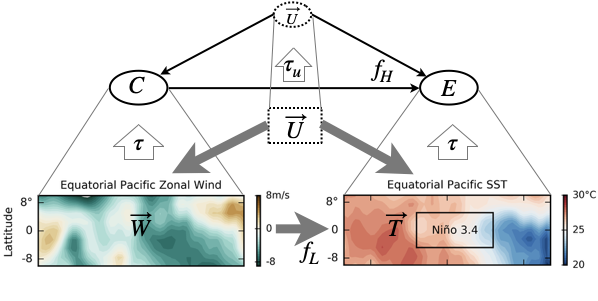}
  \caption{Climate example adapted from Chalupka et al. (2016),
    in which a high-level causal model for the phenomenon of El Ni\~no
    is constructed from low-level (high-dimensional) wind $\vec{W}$ and sea surface temperature $\vec{T}$
    measurements. $\vec{U}$ is an unmeasured confounder and $\tau$ is
        the mapping between the models. See text for details. 
  }
  \label{fig:climate} 
\end{figure}

As a simplified working example to illustrate our points we use
the case of the wind and sea surface temperature patterns over the
equatorial Pacific that give rise to the high-level climate phenomena
of El Ni\~no and La Ni\~na, as described by Chalupka et al.~\citeyear{CBPE16}.
They considered the question of how the El Ni\~no climate phenomenon
related to the underlying wind and sea surface temperature patterns
that constitute it (see Fig.~\ref{fig:climate}). At the low level they
considered two high-dimensional vector-valued variables representing
the wind speeds  $\vec{W}$ and the sea surface temperatures
$\vec{T}$, respectively, on a grid of geographical locations in the
equatorial Pacific. They assumed (with some justification
from climate science) that wind speed $\vec{W}$ is a cause
of sea surface temperature $\vec{T}$, that is, $\vec{T} =
f_L(\vec{W},\vec{U})$ for some high-dimensional function $f_L(.)$ and
exogenous causes $\vec{U}$. They allowed the possibility that
$\vec{U}$ may be a confounder of $\vec{W}$ and $\vec{T}$, so that there
might be an additional causal relation $\vec{W} =
g_L(\vec{U})$. Leaving details about feedback and temporal delay
aside, they were interested in whether the same system could be
described at a higher level, using a low-dimensional structural
equation $E = f_H(C, \vec{U})$, where there is a surjective mapping
$\tau$ from $\R(\vec{T}, \vec{W})$, the set of
possible values of $\vec{T}$ and $\vec{W}$, to $\R(E,C)$.
In the language of this paper, they were searching for an
\emph{abstract} causal description of the system.
They
required that the high-level model retain a causal interpretation, in
the sense that if one intervened on $C$, there would still be a well-defined
causal effect on $E$, no matter how the intervention on $C$ was
interpreted as an 
intervention on the underlying set of variables $\vec{W}$.

Chalupka et al.~\citeyear{CBPE16} were able to learn such a
high-level model, and one of 
the states of $E$ (the high-level description of the sea surface
temperatures) indeed corresponded to what would commonly be described
as an El Ni\~no occurring, conventionally defined by an average temperature deviation in a rectangular region of the Pacific. However, the high-level description was not
perfect: it provided an informative causal description of the
underlying systems and allowed for predictions that
\emph{approximated} the actual outcomes. Here we make
precise the nature of such an approximation between a high-level and
low-level causal model of the same system.
In the process, we define  what it means for one causal
model to approximate another.  

Although our running example is a vastly simplified climate model, the
challenge of approximately modeling phenomena at a more abstract level
is part of  almost \emph{every} scientific model. For example, it was
Robert Boyle's great insight that, despite its inaccuracies for real
gases in practice, the ideal gas law still provides an
\emph{approximate} abstract description of the behavior of the
molecules of a gas in a container that is extraordinarily useful for
understanding and manipulating real systems. Approximate abstractions
can take a variety of forms in scientific practice, ranging from
idealizations and discretizations to other forms of simplification and
dimension reduction (as in the climate example). Our account captures
these in a unified formal framework. 

\commentout{
  The main contribution of this paper is to provide what we believe are
  the ``right'' (or, at least, quite useful) definitions of
  approximation and approximate abstraction, and the related conceptual
  discussion of these notions.  In addition, we provide some technical
  results regarding the difficulty of determining whether an approximate
  abstraction can be viewed as the composition of an approximation and
  an exact abstraction.    We believe that the framework presented here
  can provide a foundation for analyzing abstraction and approximation
  in causal models.
}%
The main contribution of this paper is to present a framework that
offers a foundation for   
analyzing abstraction and approximation in causal models. We provide
what we believe are sensible definitions of approximation and
approximate abstraction, and a conceptual 
discussion of these notions. In addition, we provide some technical
results regarding the difficulty of determining whether an approximate
abstraction can be viewed as the composition of an approximation and
an exact abstraction.    


\section{PRELIMINARIES}\label{sec:review}

Since we are interested in scientific models that support explanations
of phenomena and can inform interventions on a system, we start by
defining a deterministic causal model with a set of possible
interventions. We use exogenous and
endogenous variables to distinguish those influences that are external to the
system and those that  are internal. The definitions follow the framework
developed by Halpern~\citeyear{Hal48}.


\dfn
A signature $\S$ is a tuple $(\U,\V,\R)$, where $\U$
is a set of \emph{exogenous} variables, $\V$ is a set 
of \emph{endogenous} variables,
and $\R$, a function that associates with every variable $Y \in  
\U \union \V$ a nonempty set $\R(Y)$ of possible values for $Y$
(i.e., the set of values over which $Y$ {\em ranges}).
If $\vec{X} = (X_1, \ldots, X_n)$, $\R(\vec{X})$ denotes the
crossproduct $\R(X_1) \times \cdots \times \R(X_n)$.
\edfn
For simplicity in this paper, we assume that signatures are finite,
that is, $\U$ and $\V$ are finite, and the range of each variable $Y
\in \U \union \V$ is finite.

\dfn
A \emph{basic causal model} $M$ is a pair $(\S,\F)$, 
where $\S$ is a signature and
$\F$ defines a  function that associates with each endogenous
variable $X$ a \emph{structural equation} $F_X$ giving the value of
$X$ in terms of the 
values of other endogenous and exogenous variables.
Formally, the equation $F_X$ maps $\R(\U \union \V - \{X\})$  to $\R(X)$,
so $F_X$ determines the value of $X$, 
given the values of all the other variables in $\U \union \V$.  
\edfn
Note that there are no functions associated with exogenous variables,
since their values are determined outside the model.  We call a
setting $\vec{u}$ of values of exogenous variables a \emph{context}.%
\footnote{We remark that the notion of context used here, which goes
  back to \cite{HP01b}, is similar to that of Boutilier et
  al. \citeyear{BFGK96}, in that both are assignments of values to
  variables.  However, here it is used in particular to denote an
  assignment of values to all the exogenous variables.}

The value of $X$ may not depend on the values of all other variables.  
$Y$ \emph{depends on $X$ in context $\vec{u}$}
if there is some setting of the endogenous variables  
other than $X$ and $Y$ such that 
if the exogenous variables have value $\vec{u}$, then varying the value
of $X$ in that 
context results in a variation in the value of $Y$; that 
is, there is a setting $\vec{z}$ of the endogenous variables other than $X$ and
$Y$ and values $x$ and $x'$ of $X$ such that $F_Y(x,\vec{z},\vec{u}) \ne
F_Y(x',\vec{z},\vec{u})$.

\commentout{
  In this paper we restrict attention to \emph{recursive} (or
  \emph{acyclic}) models, that is, models where, for each context
  $\vec{u}$, there is a partial order 
  $\preceq_{\vec{u}}$ on variables such that if $Y$ depends on $X$ in
  context $\vec{u}$, then $X \prec_{\vec{u}} Y$.
  In a recursive model, given a context $\vec{u}$, 
  the values of all the remaining variables are determined (we can just
  solve for the value of the endogenous variables in the order given
  by $\prec_{\vec{u}}$).  
  A model is \emph{strongly} recursive if the
  partial order $\preceq_{\vec{u}}$ is independent of $\vec{u}$; that is,
  there is a partial order $\preceq$ such that $\preceq =
  \preceq_{\vec{u}}$ for all contexts $\vec{u}$.
  (See \cite{Hal48} for more discussion of recursive vs. strongly
  recursive models.  The distinction has no impact on our results.)
  We often write the
  equation for an endogenous variable as $X = f(\vec{Y})$; this denotes
  that the value of $X$ depends only on the values of the variables in
  $\vec{Y}$, and the connection is given by $f$.  Our climate example is strongly recursive, since $\vec{T} = f_L(\vec{W},\vec{U})$. 
}%
In this paper, we restrict attention to \emph{recursive} (or
\emph{acyclic}) models, that is, models where there is a partial order 
$\preceq$ on variables such that if $Y$ depends on $X$, then $X \prec Y$.%
\footnote{Halpern \citeyear{Hal48} calls this \emph{strongly
    recursive}, in order 
  to distinguish it from models in which the 
  partial order depends on the context. This distinction has no impact on our results.}\label{note:recursive}
In a recursive model, given a context $\vec{u}$, 
the values of all the remaining variables are determined (we can just
solve for the value of the endogenous variables in the order given
by $\prec$)
We often write the
equation for an endogenous variable as $X = f(\vec{Y})$; this denotes
that the value of $X$ depends only on the values of the variables in
$\vec{Y}$, and the connection is given by $f$.  Our climate example is recursive, since $\vec{T} = f_L(\vec{W},\vec{U})$. 

An \emph{intervention} has the form $\vec{X} \gets \vec{x}$, where $\vec{X}$ 
is a set of endogenous variables.  Intuitively, this means that the
values of the variables in $\vec{X}$ are set to $\vec{x}$.  
The structural equations define what happens in the presence of 
interventions.  Setting the value of some variables $\vec{X}$ to
$\vec{x}$ in a causal 
model $M = (\S,\F)$ results in a new causal model, denoted $M_{\vec{X}
  \gets \vec{x}}$, which is identical to $M$, except that $\F$ is
replaced by $\F^{\vec{X} \gets \vec{x}}$: for each variable $Y \notin
\vec{X}$, $F^{\vec{X} \gets \vec{x}}_Y = F_Y$\fullv{
(i.e., the equation for $Y$ is unchanged)}, while for
each $X'$ in $\vec{X}$, the equation $F_{X'}$ for is replaced by $X' = x'$
(where $x'$ is the value in $\vec{x}$ corresponding to $\vec{x}$).

Halpern and Pearl \citeyear{HP01b} and Halpern \citeyear{Hal48} 
implicitly assumed that all 
interventions can be performed in a model. For reasons that will
become clear when defining abstraction, we follow Rubenstein et
al.~\citeyear{Rub17} and  Beckers and Halpern \citeyear{BH19} in adding
the notion of ``allowed interventions'' to a causal model. This allows
us to capture situations where not all interventions are of interest to the
modeler and/or some interventions may not be feasible.  We can
then define a  \emph{causal model} $M$ as a tuple $(\S,\F,\I)$, where
$(\S,\F)$ is a basic causal model and $\I$ is a set of \emph{allowed
  interventions}.
We sometimes write a causal model $M = (\S,\F,\I)$ as $(M',\I)$, where 
$M'$ is the basic causal model $(\S,\F)$, if we want to emphasize the
role of the allowed interventions.

Given a signature $\S = (\U,\V,\R)$, a \emph{primitive event} is a
formula of the form $X = x$, for  $X \in \V$ and $x \in \R(X)$.  
A {\em causal formula (over $\S$)\/} is one of the form
\mbox{$[Y_1 \gets y_1, \ldots, Y_k \gets y_k] \phi$},
where
$\phi$ is a Boolean
combination of primitive events, 
$Y_1, \ldots, Y_k$ are distinct variables in $\V$, and
$y_i \in \R(Y_i)$.
Such a formula is
abbreviated
as $[\vec{Y} \gets \vec{y}]\phi$.
The special
case where $k=0$
is abbreviated as
$\phi$.
Intuitively,
$[Y_1 \gets y_1, \ldots, Y_k \gets y_k] \phi$ says that
$\phi$ would hold if
$Y_i$ were set to $y_i$, for $i = 1,\ldots,k$.

A causal formula $\psi$ is true or false in a causal model, given a
context. As usual, we write $(M,\vec{u}) \sat \psi$  if the causal
formula $\psi$ is true in causal model $M$ given context $\vec{u}$.
The $\sat$ relation is defined inductively.
$(M,\vec{u}) \sat X = x$ if the variable $X$ has value $x$
in the unique (since we are dealing with recursive models) solution
to the equations in $M$ in context $\vec{u}$ (i.e., the  unique vector
of values that simultaneously satisfies all
equations in $M$ with the variables in $\U$ set to $\vec{u}$).
The truth of conjunctions and negations is defined in the standard way.
Finally, $(M,\vec{u}) \sat [\vec{Y} \gets \vec{y}]\phi$ if 
$(M_{\vec{Y} = \vec{y}},\vec{u}) \sat \phi$.

To simplify notation, we sometimes write $M(\vec{u})$ to denote the
unique element of $\R(\V)$ such that $(M,\vec{u}) \sat \V = \vec{v}$.
Similarly, given an intervention $\vec{Y} \gets \vec{y}$, 
$M(\vec{u},\vec{Y} \gets \vec{y})$
denotes the unique element of $\R(\V)$ such that $(M,\vec{u}) \sat
[\vec{Y} \gets \vec{y}](\V = \vec{v})$.

\commentout{
  A \emph{probabilistic causal model} $M=(\cal S,\F,\I,\Pr)$ is just a
  causal model together with a probability $\Pr$ on contexts.  
  We often abuse notation slightly and denote the probabilistic causal
  model $(\cal S,\F,\I,\Pr)$ as $(M,\Pr)$, where $M$ is the underlying
  deterministic causal model $(\cal S,\F,\I)$.
}%

These definitions allow us to describe the climate system both in
terms of a low-level causal model $M_L$ and a high-level model $M_H$: 
\begin{align*}
  M_L &= ((\U_L,\V_L,\R_L), \F_L,\I_L)\\
  &= ((\{\vec{U}\}, \{\vec{W}, \vec{T}\}, \R_L), \{g_L, f_L\}, \I_L)
  \mbox{ and }\\
  M_H &= ((\U_H,\V_H,\R_H), \F_H,\I_H)\\
  &= ((\{\vec{U}\}, \{C, E \}, \R_H), \{g_H, f_H\}, \I_H).
\end{align*}
Chalupka et al.~\citeyear{CBPE16} treated $\vec{U}$ as not only
exogenous, but also 
unobserved, and therefore made no claim about its dimensionality in the
high-level model.
We do not explicitly
spell out $\R_L$ and $\R_H$ here; the description of the climate
model indicates that $\R_L(C) \times \R_L(E)$ is much smaller than
$\R_L(\vec{W}) \times \R_L(\vec{T})$:
many different low-level (vector-valued) states may
correspond to one high-level low-dimensional state. Finally, for our
climate example we have not yet said anything about interventions, so
$\I_L$ and $\I_H$ are currently placeholders. 

\fullv{
We are now in a position to specify a relation between the high and
low-level models $M_H$ and $M_L$.
}

\subsection{Abstraction}

Beckers and Halpern \citeyear{BH19} gave a sequence of successively
more 
restrictive definitions
of abstraction for causal model.  The first and least
restrictive definition is the notion of \emph{exact transformation}
due to Rubenstein et al. \citeyear{Rub17}.  Examples given by Beckers and
Halpern show that the notion of exact transformation is arguably too
flexible.  
\commentout{Thus, in this paper we focus on the more restrictive notion
  of \emph{$\tau$-abstraction}, since this seems most appropriate for
  our purposes here, although our basic approach should be applicable to
  all the notions of abstraction considered by Beckers and Halpern.  
}
Thus, in this paper the notion of abstraction we consider is that
of \emph{$\tau$-abstractions},
introduced by Beckers and Halpern, which can be viewed
as a restriction of exact transformations%
\commentout{
  \footnote{Exact $\tau$-transformations relate probabilistic
    causal models, while $\tau$-abstractions relate (deterministic) causal
    models.  We can compare the two by showing that every
    $\tau$-abstraction is what Beckers and Halpern \citeyear{BH19} call a
    \emph{uniform $\tau$-transformation}: specifically, if $M_H$ is a
    $\tau$-abstraction of $M_L$, then for every probability $\Pr_L$, there
    exists a probability $\Pr_H$ such that $(M_L,\Pr_L)$ is an exact
    $\tau$-transformation of $(M_H,\Pr_H)$.  This is shown in the full version of \cite{BH19}.
  } 
}%
\footnote{Exact $\tau$-transformations relate probabilistic
  causal models, while $\tau$-abstractions relate (deterministic) causal
  models. Beckers and Halpern \citeyear{BH19} show that we can compare the two by proving that every
  $\tau$-abstraction is what they call a
  \emph{uniform $\tau$-transformation}: specifically, if $M_H$ is a
  $\tau$-abstraction of $M_L$, then for every probability $\Pr_L$, there
  exists a probability $\Pr_H$ such that $(M_L,\Pr_L)$ is an exact
  $\tau$-transformation of $(M_H,\Pr_H)$.
} 
and avoids some of their problems.
However, 
nothing hinges on this choice: all definitions can just as well
be interpreted using the less restrictive notions of
abstractions.

The key to defining all the notions of abstraction from a low-level to
a high-level causal model considered by Beckers and Halpern is the
abstraction function \mbox{$\tau: \R_L(\V_L) \rightarrow \R_H(\V_H)$}
that maps endogenous states of $M_L$ to endogenous states of $M_H$.
This is a generalization of the surjective mapping $\tau$ from
$\R(\vec{T},\vec{W})$ to $\R(E,C)$ discussed in the introduction.
In the formal definition, we need two additional functions: \mbox{$\tau_{\U}:
  \R_L(\U_L) \rightarrow \R_H(\U_H)$}, which maps exogenous states of $M_L$
to exogenous states of $M_H$, and \mbox{$\omega_{\tau}: \I_L \rightarrow
  \I_H$}, which  maps low-level interventions to high-level interventions.
Beckers and Halpern~\citeyear{BH19} show that, given their definition
of abstraction, $\tau_{\U}$ and $\omega_\tau$ can be derived from $\tau$.
We briefly review the relevant definitions here; we refer the reader
to their paper for 
more details and 
motivation. 

\dfn\label{def:induce}
Given a set $\V$ of endogenous variables, $\vec{X} \subseteq \V$, and $\vec{x}
\in \R(\vec{X})$, let 
$$\Rest(\V,\vec{x}) = \{\vec{v} \in \R(\V): \vec{x}
\mbox{ is the restriction of } \vec{v} \mbox{ to } \vec{X}\}.$$
Given $\tau: \R_L(\V_L) \rightarrow \R_H(\V_H)$, 
define $\omega_{\tau}(\vec{X} \gets \vec{x}) = \vec{Y} \gets \vec{y}$ if
$\vec{Y} \subseteq \V_H$, $\vec{y} \in \R_H(\vec{Y})$, and
$\tau(\Rest(\V_L,\vec{x})) = \Rest(\V_H,\vec{y})$ (where, as usual, given
$T \subseteq \R_L(\V_L)$, define $\tau(T) = \{\tau(\vec{v}_L):
\vec{v}_L \in T\}$).
It is easy to see
that, given $\vec{X}$ and $\vec{x}$, there can be at most one such
$\vec{Y}$ and $\vec{y}$.
If  such a $\vec{Y}$ and $\vec{y}$ do not exist,
we take \mbox{$\omega_{\tau}(\vec{X} \gets \vec{x})$} to be
undefined.
Let $\I^\tau_L$ be the set of interventions for which
$\omega_{\tau}$ is defined, and let $\I^\tau_H = \omega_{\tau}(\I^\tau_L)$.  
\edfn

Note that if  $\tau$ is surjective, then it easily
follows that
$\omega_{\tau}(\emptyset)=\emptyset$, 
and for all $\vec{v}_L \in \R_L(\V_L)$, $\omega_{\tau}(\V_L \gets
\vec{v}_L)=\V_H \gets \tau(\vec{v}_L)$. 

With this definition, the need for the intervention sets $\I_L$ and
$\I_H$ becomes clear: in general, not all low-level interventions will
neatly map to a high-level intervention, since the abstraction
function $\tau$ may aggregate variables together; some
low-level interventions will constitute only a partial intervention on
a high-level variable. The ``allowed intervention'' sets ensure that
the set of interventions can be suitably restricted to retain only
those that can actually be abstracted. Similarly, there may be cases
where the high-level model does not support all interventions because
they may not be well-defined. For example, what does it mean in the
ideal gas law to change temperature, while keeping pressure and volume
constant?
It is not even clear that such an intervention is meaningful.

Of course, a minimal requirement for any causal model to be a
$\tau$-abstraction of some other model is that the signatures of both
models need to be compatible with $\tau$. Beckers and Halpern~\citeyear{BH19} add
two further minimal requirements.  We capture all
of them
by requiring 
the two causal models to be $\tau$-consistent: 
\dfn
If $\tau: \R_L(\V_L) \rightarrow \R_H(\V_H)$, then
$(M_H, \I_H)$ and $(M_L, \I_L)$ are \emph{$\tau$-consistent} if 
$\tau$ is surjective,
$\I_H = \omega_{\tau}(\I_L)$, and
$ |\R_L(\U_L)| \ge |\R_H(\U_H)|$.
\edfn


\dfn\label{dfn:tau-abstraction}
$(M_H, \I_H)$ is a \emph{$\tau$-abstraction} of $(M_L, \I_L)$ if
$(M_H, \I_H)$ and $(M_L, \I_L)$ are $\tau$-consistent and there exists
a surjective $\tau_{\U}$ such that
for all $\vec{u}_L \in \R_L(\U_L)$ and $\vec{X} \gets \vec{x} \in \I_L$,
$\tau(M_L(\vec{u}_L,\vec{X}  \gets \vec{x})) =
M_H(\tau_{\U}(\vec{u}_L),\omega_{\tau}(\vec{X} \gets \vec{x})).$
\edfn
\commentout{
  \dfn If $(M_H, \I_H)$ and $(M_L, \I_L)$
  are both $\tau$- and $\tau_{\U}$-consistent, then 
  $\tau_{\U}$ is 
  \emph{compatible} with $\tau$ if,
  for all $\vec{X} \gets \vec{x} \in \I_L$ and $\vec{u}_L \in \R_L(\U_L)$,
  $$\tau(M_L(\vec{u}_L,\vec{X}  \gets \vec{x})) =
  M_H(\tau_{\U}(\vec{u}_L),\omega_{\tau}(\vec{X} \gets \vec{x})).$$ 
  \edfn
}%
\commentout{
  \dfn We say that $(M_H, \I_H)$ and $(M_L, \I_L)$
  are \emph{compatible} with a given $\tau$, $\tau_{\U}$ and $\omega_{\tau}$ if the following are satisfied:
  \begin{itemize}
  \item $\tau: \R_L(\V_L) \rightarrow \R_H(\V_H)$ is surjective,
  \item $\I_H = \omega_{\tau}(\I_L)$
  \item $\tau_\U: \R_L(\U_L) \rightarrow \R_H(\U_H)$ is surjective, and
  \item for all $\vec{X} \gets \vec{x} \in \I_L$ and $\vec{u}_L \in \R_L(\U_L)$,
    $$\tau(M_L(\vec{u}_L,\vec{X}  \gets \vec{x})) =
    M_H(\tau_{\U}(\vec{u}_L),\omega_{\tau}(\vec{X} \gets \vec{x})).$$ 
  \end{itemize}
  \edfn
}
\commentout{
  \dfn\label{dfn:tau-abstraction}
  $(M_H, \I_H)$ is a \emph{$\tau$-abstraction} of $(M_L, \I_L)$ if
  $(M_H, \I_H)$ and $(M_L, \I_L)$ are $\tau$-consistent and there exists a $\tau_{\U}$ that is compatible with $\tau$.
  \edfn
}%

\commentout{
  \dfn\label{def:strong} If $M_H$ and $M_L$ are basic causal models, then $M_H$ is a \emph{strong $\tau$-abstraction} of $M_L$ if  
  $(M_H,\I_H^*)$ is a $\tau$-abstraction of
  $(M_L,\I_L^{\tau})$ (where $\I_H^*$ is the set of all
  high-level interventions). 
  \edfn
}

Abstraction means that for each possible low-level
context-intervention pair, the two ways of moving up ``diagonally'' to a
high-level endogenous state always lead to the same result. The first
way is to start by applying $M_L$ to get a low-level 
state, and then moving up to a high-level state by applying
$\tau$, whereas the second way is to first move to a high-level context and
intervention (by applying $\tau_{\U}$ and $\omega_{\tau}$), 
and then to obtain a high-level
state by applying $M_H$. 

A common and useful form of abstraction occurs when the
low-level variables are clustered, so that the clusters form the
high-level variables.
Roughly speaking, the intuition
is that in the
high-level model, one variable captures the effect of a number of
variables in the low-level model.  This makes sense only if the low-level
variables that are being clustered together ``work 
the same way'' as far as the 
allowed
interventions go.  
The following definition makes this 
special case of abstraction
precise.

\dfn\label{dfn:constructive}
If $\V_H = \{Y_1, \ldots, Y_n\}$, then
$\tau: \R_L(\V_L) \rightarrow \R_H(\V_H)$ is \emph{constructive}
if there exists 
a partition $P = \{\vec{Z}_1, \ldots, \vec{Z}_{n+1}\}$ of $\V_L$,
where $\vec{Z}_1, \ldots, \vec{Z}_n$ are nonempty,
and
mappings $\tau_i: \R(\vec{Z}_i) \rightarrow \R(Y_i)$ for $i=1, \ldots, n$
such that $\tau = (\tau_1, \ldots, \tau_n)$; that is,
$\tau(\vec{v}_L) = \tau_1(\vec{z}_1)\cdot \ldots \cdot
\tau_n(\vec{z}_n)$, where $\vec{z}_i$ is the projection of $\vec{v}_L$
onto the variables in $\vec{Z}_i$, and $\cdot$ is the concatenation
operator on sequences.
\commentout{
  $M_H$ is a \emph{constructive
    abstraction} of $M_L$ if it is a
  $\tau$-abstraction of $M_L$ for some constructive $\tau$
  and $\I_L=\I_L^{\tau}$.
}%
If $M_H$ is a $\tau$-abstraction of $M_L$ then we say it is \emph{constructive} if $\tau$ is constructive and $\I_L=\I_L^{\tau}$.
\edfn

In this definition, we can think of each $\vec{Z}_i$ as describing a
set of microvariables that are mapped to a single macrovariable $Y_i$.
The variables in $\vec{Z}_{n+1}$ (which might be empty) are ones that are
marginalized away.

\begin{figure}
  \includegraphics[width = 3.5cm]{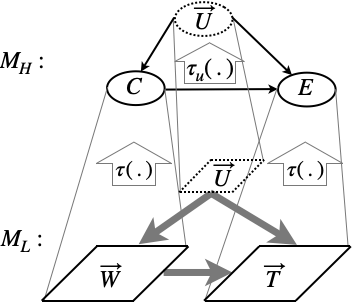}
  \hspace{4mm}
    \includegraphics[width = 3.5cm]{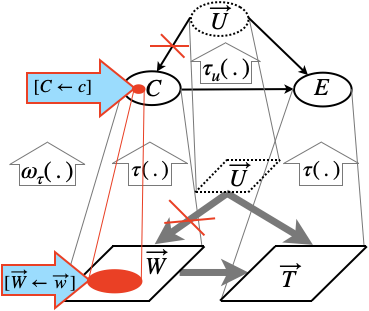}
  \caption{The climate model exemplifies constructive abstraction.
    \emph{Left:} 
    $\tau$ and $\tau_{\U}$
    map low-level variables to their high-level
    counterparts. \emph{Right:}
    $\omega_{\tau}$ maps low-level
    interventions to the high-level intervention.
    Several
    different low-level interventions on $\vec{W}$ may correspond to the
    same intervention on $C$.}
  \label{fig:constr_abstr} 
\end{figure}

The climate example almost exactly fits the notion of constructive
abstraction: the variables in the high-level model, $C$ and $E$, each
correspond to a vector-valued low-level variable, $\vec{W}$ and
$\vec{T}$, but $\vec{W}$ and $\vec{T}$ could have been replaced by 
disjoint sets of variables.
Consequently,  $\tau$ maps states from the same
low-level variable to the same high-level variable (see Fig.~\ref{fig:constr_abstr}, left). Although
interventions are practically not feasible in the climate case,
hypothetically they are perfectly well-defined: an intervention on $C$
can be instantiated at the low level by several different
interventions on $\vec{W}$ (see Fig.~\ref{fig:constr_abstr}, right). Finally, given an intervention on
$\vec{W}$, we have that 
\begin{align*}
  M_L(&\vec{u}, \vec{W} \gets \vec{w}) \sat (\vec{T} = \vec{t})
  \mbox{ iff }\\
  M_H(&\tau_{\U}(\vec{u}), \omega_{\tau}(\vec{W}
  \gets \vec{w})) \sat (E = \tau(\vec{t})).
\end{align*} 
The correspondence between
$M_L$ and $M_H$ is \emph{exact}. In fact, the 
high-level model $M_H$ constructed by Chalupka et al. \citeyear{CBPE16}
did not satisfy this biconditional precisely, but had to approximate
it. We maintain 
that, in general, high-level models in science are only \emph{approximate
  abstractions}.  

\section{APPROXIMATE ABSTRACTION}\label{sec:approximation}

In order to define what it means for one causal model to be an
approximation of another, we need a way of measuring the ``distance''
between causal models.   We take a \emph{distance function} to simply be a
function that associates with a pair $(M_1,M_2)$ a distance, that is,
a non-negative real number.
We show how various distance functions on causal models can be
defined, starting from a 
\emph{metric} $d_{\V}$  on the state space $\R(\V)$
of a causal model.  (Recall that a metric  
on a space $X$ is a function $d: X \times X \rightarrow \IR^+$ such that
(a) $d(x,x') = 0$ iff $x = x'$, (b) $d(x,y) = d(y,x)$,  and (c)
$d(x,y) + d(y,z) \ge d(x,z)$.) 
Such a metric $d_{\V}$ is typically
straightforward to define.  Given two states $s_1$ and $s_2$, we can
compare the value of each endogenous variable $X$ in $s_1$ and $s_2$.
The difference in the values determines the distance between $s_1$ and
$s_2$.

The choice of distance function is application-dependent.  
Different researchers looking at the same data may be interested in different
aspects of the data. For example, suppose that the model is defined in
terms of 5 variables, $X_1, \ldots, X_5$. $X_3$ might be gender and $X_4$ might
be height. Suppose that we restrict to distance functions that takes
the distance between $(x_1, \ldots, x_5)$ and $(x_1', ..., x_5')$ to be of
the form $\sqrt{w_1(x_1 - x_1')^2 + \cdots + w_5(x_5 - x_5')^2}$, where
$w_1, \ldots, w_5$ are weights that represent the importance (to the
researcher) of each of these five features. One researcher might not be
interested in gender (so doesn't care if her predictions about gender
are incorrect), and thus might take $w_3 = 0$; another researcher might
care about gender and not about height, so she might take $w_3 = 1$ and
$w_4 = 0$.  While, as we shall see, the choice of distance function
makes a crucial difference in evaluating the ``goodness'' of an
approximate abstraction, in light of the above, we leave the choice of
distance function unspecified.

In the remainder of the paper, we assume that the state space
$\R(\V)$ of endogenous variables for each causal model comes with a
metric $d_{\V}$.
We provide a number of ways of lifting the metric $d_{\V}$ on states
to a distance function $d$ on models,
and then use the distance function to define both approximation and
approximate abstraction.

Our intuition for the distance function is based on 
how causal models are typically used.  Specifically, we are interested
in
how two models compare with regard to the predictions they make about the  
effects of an intervention.
Our intuition is similar in spirit to that behind the notion of
\emph{structural intervention distance} considered by Peters and B\"uhlmann
\citeyear{PB15}, although the technical definitions are quite
different.  (We discuss the exact relationship between our 
approach and theirs in the next section, in the context of 
probabilistic models.) 

We start with the simplest setting where this intuition can be made
precise, one where the models $M_1$ and $M_2$ differ only with regards
to their equations.
We say that two models are \emph{similar} in this case.
If two models are similar, then, among other things, we can assume
that they have the same metric $d_{\V}$.
In this setting, we can compare the effect of each allowed
intervention $\vec{X} \gets \vec{x}$ in the two models.
That is, for each context $\vec{u}$, we can compare the states
$M_1(\vec{u},\vec{X}  \gets  \vec{x})$ and $M_2(\vec{u},\vec{X}  \gets
\vec{x})$ that arise after performing the intervention $\vec{X}
\gets \vec{x}$ in context $\vec{u}$ in each model.  We get the desired
distance function  by taking the worst-case distance between all such states.

\dfn\label{def:max}
\commentout{
  Fix a signature $\S = (\U,\V,\R)$ and a set $\I$ of interventions, and let
  $\M^{(\S,\I)}$ consist of all causal models with signature $\S$
  and set $\I$ of interventions (so that models in $\M^{(\S,\I)}$
  differ only in their equations).  Define a 
  distance function $d_{max}^{(\S,\I)}$ on $\M^{(\S,\I)}
  \times 
  \M^{(\S,\I)}$ by taking 
}%
Define a distance function $d_{max}$ on pairs of 
similar models by taking
$$\begin{array}{ll}
  d_{max}(M_1,M_2)=\\
  \max_{\vec{X} \gets \vec{x} \in \I \text{, }
    \vec{u} \in \R(\U)} \{  d_{\V}(M_1(\vec{u},\vec{X} \gets
  \vec{x}),M_2(\vec{u},\vec{X}  \gets  \vec{x}))\}.%
  \commentout{
    \footnote{\protect{We remark that $d_{max}^{Id}$ is a \emph{pseudometric}: it
        satisfies the second and third properties of a metric, but we may
        have $d_{max}^{Id}(M_1,M_2) = 0$ even if $M_1 \ne M_2$.
        But if, $\I=\I^*$  (the set of all interventions), then
        $d_{max}^{(\U,V,\R,),\I)}$    is a metric.}}
  }
\end{array}
$$ 
The causal model $M_1$ is a \emph{$d_{max}$-$\alpha$ approximation} of $M_2$
if $d_{max}(M_1,M_2) \le \alpha$.
\edfn
Thus, $M_1$ is a $d_{max}$-$\alpha$ approximation of $M_2$ if the
predictions of $M_1$ are always within $\alpha$ of the predictions of $M_2$.

We apply similar ideas to defining approximate abstraction.
But now we no longer have a distance function defined on
causal models with 
the same signature.  Rather, the distance function $d_{\tau}$ is
defined on pairs $(M_L,M_H)$ consisting of a low-level and high-level 
causal model (which, in general, have different signatures), related by a 
a surjective mapping $\tau$.
The idea behind $d_\tau$ is that we start with a low-level intervention
$\vec{X} \gets \vec{x}$, consider its effects in $M_L$, lift this up
to $M_H$  using $\tau$, and compare this to the effects of
$\omega_\tau(\vec{X} \gets \vec{x})$ in $M_H$.

\dfn
\commentout{
  Fix a surjective map $\tau: \R(\V_L) \rightarrow \R(\V_H)$ 
  and $\omega_\tau(\I_L) = \I_H$.
  Then the distance function 
  $d_{\tau}$
  is defined on pairs $(M_L,M_H)$ such that  
  $M_L$ has signature $(\U_L,\V_L,\R_L)$ and interventions $\I_L$ and
  $M_H$ has signature $(\U_H,\V_H,\R_H)$ and interventions $\I_H$ by taking
}%
Fix a surjective map $\tau: \R(\V_L) \rightarrow \R(\V_H)$. Define the distance function 
$d_{\tau}$ on pairs of $\tau$-consistent models $(M_L,M_H)$ by taking
\commentout{
  $$\begin{array}{ll}
    d_{\tau}(M_L,M_H)= \\\min_
    {\mathrm{\tau_{\U} \ surjective}}
    \max_{\vec{X} \gets \vec{x} \in \I_L \text{, }
      \vec{u}_L \in \R_L(\U_L)}\\ ( d_{\V_H}(\tau(M_L(\vec{u}_L,\vec{X} \gets
    \vec{x})),M_H(\tau_{\U}(\vec{u}_L),\omega_{\tau}(\vec{X} \gets
    \vec{x}) ))).\end{array}$$  
}%
$$\begin{array}{ll}
  d_{\tau}(M_L,M_H)= \min_
  {\mathrm{\tau_{\U} \ surjective}}
  \max_{\vec{X} \gets \vec{x} \in \I_L \text{, }
    \vec{u}_L \in \R_L(\U_L)}\\ ( d_{\V_H}(\tau(M_L(\vec{u}_L,\vec{X} \gets
  \vec{x})),M_H(\tau_{\U}(\vec{u}_L),\omega_{\tau}(\vec{X} \gets
  \vec{x}) ))).\end{array}$$  
$M_H$ is a \emph{$\tau$-$\alpha$ 
  approximate
  abstraction} of $M_L$ if
$d_\tau(M_L,M_H) \le \alpha$.
\commentout{
  \emph{A constructive $\tau$-$\alpha$ approximate abstraction} is a 
  $\tau$-$\alpha$ approximate abstraction where $\tau$ is constructive
  and $\I_L = \I_L^{\tau}$.
}%
\edfn
We take the minimum over all functions $\tau_\U$ because
the function that lifts the low-level contexts up to the high-level
contexts does not play a major role. We thus simply focus on the best
choice of $\tau_\U$.  

\begin{figure}
  \centering
  \includegraphics[width = 4.5cm]{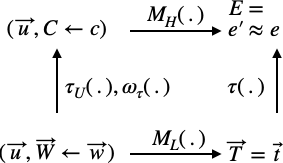}
  \caption{Approximate abstraction in the climate example.  We 
    measure (for the worst-case
    choice of low-level intervention $\vec{W} \gets \vec{w}$ and context  $\vec{u}$)
    the distance between the values of the variable $E$ obtained by (a)
    first applying $M_L$ to 
    $\vec{u}$ and $\vec{W} \gets \vec{w}$ and then abstracting by
    applying $\tau$ vs.\ (b)  abstracting the intervention and context
    (by applying $\omega_\tau$ and $\tau_\U$, respectively) and then
    applying $M_H$.} 
  \label{fig:approxabstr} 
\end{figure}

To get an intuition for an approximate abstraction,
consider the climate example again. For a low-level intervention
$\vec{W} \gets \vec{w}$ 
and a low-level context $\vec{u}$, there are
two ways of lifting 
their effect to $M_H$ (see Fig.~\ref{fig:approxabstr}). The first is to  
start by applying $M_L$ to the context-intervention pair to
determine a low-level state $\vec{t}$, and then 
apply $\tau$ to obtain the high-level state $E = e = \tau(\vec{t})$.
(Recall that  $M_L$ can be viewed as 
a function from context-intervention pairs to states in $\R_L(\V_L)$.)
The second is to first lift the intervention  $\vec{W} \gets \vec{w}$ to $\I_H$, that is, an
intervention on $C$, 
and the context to $\U_H$ by applying  $\omega_\tau$ and
$\tau_{\U}$. Then we apply $M_H(\vec{u}_H, C \gets c)$, which again gives a high-level endogenous
state $e'$. 
We are identifying the degree to which $M_H$ approximates $M_L$ 
with
 the worst-case
distance between the two ways of lifting the context-intervention
pairs, for an optimal choice of $\tau_{\U}$. 

The following straightforward results show that our notion of
approximate abstraction is a sensible generalization of both the
notion of 
an exact
abstraction and
the notion of approximation between
similar models.

\pro
$M_H$ is a $\tau$-$0$-approximate abstraction of $M_L$ iff  $M_H$ is a $\tau$-abstraction of $M_L$. 
\epro

\pro
%
If $M_1$ and $M_2$ are 
similar, then
$M_2$ is a $\mathit{Id}$-$\alpha$-approximate abstraction of $M_1$
(where $\mathit{Id}$ is the identity function on $\R(\V)$)
iff  $M_2$ is a
$d_{max}$-$\alpha$-approximation of $M_1$.
\epro

\section{APPROXIMATE ABSTRACTION FOR PROBABILISTIC CAUSAL MODELS}


A \emph{probabilistic causal model} $M=(\cal S,\F,\I,\Pr)$ is just a
causal model together with a probability $\Pr$ on contexts $\vec{u}$.  
\commentout{ 
  We often abuse notation slightly and denote the probabilistic causal
  model $(\cal S,\F,\I,\Pr)$ as $(M,\Pr)$, where $M$ is the underlying
  deterministic causal model $(\cal S,\F,\I)$.}
\commentout{
  First, we extend the notion of $\tau$-consistency to probabilistic causal models. 
  \dfn
  Given $\tau:  \A \rightarrow \B$, $(M_H, \I_H,\Pr_H)$ and $(M_L, \I_L,\Pr_L)$ are \emph{$\tau$-consistent}  if either
  the following conditions hold:
  \begin{itemize}
  \item $\tau$ is surjective;
  \item $\A=\R_L(\V_L)$ and $\B=\R_H(\V_H)$;
  \item $\I_H = \omega_{\tau}(\I_L)$.
  \end{itemize}

  Or if the following conditions hold:
  \begin{itemize}
  \item $\tau$ is surjective;
  \item $\A=\R_L(\U_L)$ and $\B=\R_H(\U_H)$;
  \item $\tau(\Pr_L)=\Pr_H$.
  \end{itemize}
  \edfn
}%
In this section, we assume that all causal models are probabilistic,
and extend the notion of approximation to 
probabilistic causal models.  
We again start by considering the simplest setting, where we have 
probabilistic models that differ only in their equations.
We again call such models \emph{similar}.
Now we have several reasonable distance functions.  

\dfn
\commentout{
  Let $\M^{(\S,\I,\Pr)}$ consist of all probabilistic causal
  models with signature $\S = (\U,\V,\R)$, set $\I$ of interventions, and
  probability 
  $\Pr$ on contexts (so, again, that models in $\M^{(\S,\I,\Pr)}$
  differ only in their equations).  Define the
  distance function $d_{max}^{(\S,\I,\Pr)}$
  on $\M^{(\S,\I,\Pr)} \times   \M^{(\S,\I,\Pr)}$
  by taking
}%
Define a distance function $d_{max}$ on pairs of similar probabilistic causal models by taking
$$\begin{array}{ll}
  d_{max}(M_1,M_2)=
  \max_{\vec{X} \gets \vec{x} \in \I}\\
  \left(\sum_{\vec{u} \in \R(\U)} \Pr(\vec{u}) d_{\V}(M_1(\vec{u},\vec{X} \gets
  \vec{x}),M_2(\vec{u},\vec{X}  \gets  \vec{x}))\right).\end{array}$$
The probabilistic causal model $M_1$ is a
\emph{$d_{max}$-$\alpha$
  approximation} of $M_2$ 
if $d_{max}(M_1,M_2) \le \alpha$.
\edfn
Here
we have just replaced the max over contexts in
Definition~\ref{def:max} by an expectation over contexts.

But we may not always just be interested in the expected distance.  We
may, for example, be more concerned with the likelihood of serious
prediction differences, and not be too concerned about small
differences.  This leads to the following definition.

\dfn\label{def:beta}  Define a distance function $d_{\beta}$ on
pairs of similar probabilistic causal models by taking 
$$\begin{array}{ll}
  d_{\beta}(M_1,M_2)= \max_{\vec{X} \gets \vec{x} \in \I}\\
  \Pr(\{\vec{u}: d_{\V}(M_1(\vec{u},\vec{X} \gets
  \vec{x}),M_2(\vec{u},\vec{X}  \gets  \vec{x})) \ge \beta\}.
\end{array}$$
The probabilistic causal model $M_1$ is a \emph{$d_{\beta}$-$\alpha$
  approximation} of $M_2$ 
if $d_{\beta}(M_1,M_2) \le \alpha$.
\edfn

We can now extend these ideas to approximate abstraction.  We first
extend the definition of $\tau$-abstraction to the probabilistic setting.

Note that we can view 
$\Pr$ as a probability measure on $\R(\V)$, by taking $\Pr(\vec{v})
= \{\vec{u}: M(\vec{u}) =\vec{v}\}$.
An intervention $\vec{X} \gets \vec{x}$ also induces a probability
$\Pr^{\vec{X} \gets \vec{x}}$ on $\R(\V)$ in the obvious way:
$${\Pr}^{\vec{X} \gets \vec{x}}(\vec{v})
= \Pr(\{\vec{u}: M(\vec{u},\vec{X} \gets \vec{x})= \vec{v}\}).$$
In the deterministic notion of abstraction, we require that the two
high-level states obtained by the two different ways of lifting the
effects of a low-level intervention to the high level be equal.  In the
probabilistic notion, we require that the two different probability
distributions obtained by the two ways of lifting an intervention be
equal.  

\dfn\label{dfn:prob-tau-abstraction}
$M_H$ is a \emph{$\tau$-abstraction} of $M_L$ if
\commentout{
  the following conditions hold:
  \begin{itemize}
  \item $\tau$ is surjective;
  \item  $\I_H = \omega_{\tau}(\I_L)$;
  \item there is a surjective function
    $\tau_{\U}: \R_L(\U_L) \rightarrow \R_H(\U_H)$ such that $\Pr_H=\tau_{\U}({\Pr}_L)$;
  \item for all interventions $\vec{X} \gets \vec{x} \in \I_L$, we have
    that $\tau({\Pr}_L^{\vec{X} \gets \vec{x}}) =
    {\Pr}_H^{\omega_{\tau}(\vec{X} \gets \vec{x})}.$ 
    \ \  \bbox
  \end{itemize}
}%
$M_H$ and $M_L$ are $\tau$-consistent 
and
for all interventions $\vec{X} \gets \vec{x} \in \I_L$, we have
that $\tau({\Pr}_L^{\vec{X} \gets \vec{x}}) =
{\Pr}_H^{\omega_{\tau}(\vec{X} \gets \vec{x})}.$ 
\edfn

We can now extend our definitions to the 
approximate
scenario 
just as we did for deterministic causal models. 

\dfn\label{def:dist-prob}
\commentout{
  Fix a surjective map $\tau: \R(\V_L) \rightarrow \R(\V_H)$, 
  $\omega_\tau(\I_L) = \I_H$ 
  and $\tau_{\U}(\Pr_L)=\Pr_H$.
  Then the distance function 
  $d_{\tau}^{(\Pr_L,\Pr_H)}$ is defined on
  pairs of probabilistic causal models 
  $(M_L,M_H)$ such that  
  $M_L$ has signature $(\U_L,\V_L,\R_L)$,
  interventions $\I_L$, and distribution $\Pr_L$, 
  and $M_H$ has signature $(\U_H,\V_H,\R_H)$, interventions $\I_H$ and distribution $\Pr_H$ by taking
}%
Fix a surjective map $\tau: \R(\V_L) \rightarrow \R(\V_H)$. Define the distance function $d_{\tau}$ on pairs of $\tau$-consistent 
probabilistic causal models by taking
$$\begin{array}{ll}
  d_{\tau}(M_L,M_H)= 
  \min_{\mathrm{\{\tau_{\U}:
      \tau_{\U}(\Pr_L)=\Pr_H\}}}
  \max_{\vec{X} \gets \vec{x} \in \I_L} \\
  (
  \sum_{\vec{u}_L \in \R_L(\U_L)} \Pr_L(\vec{u}_L) \\
  d_{\V_H}(\tau(M_L(\vec{u}_L,\vec{X} \gets
  \vec{x})),M_H(\tau_{\U}(\vec{u}_L),\omega_{\tau}(\vec{X}  \gets  \vec{x})))
  ).
\end{array}$$
$M_H$ is a \emph{$\tau$-$\alpha$ approximate
  abstraction} of $M_L$ if $d_\tau(M_L,M_H) \le \alpha$.
\edfn
For the climate example this definition implies the following: Suppose
we introduce probabilities by specifying distributions over the
contexts $\vec{u}_L$ and $\vec{u}_H$. The resulting
probabilistic causal model
$(M_H, Pr_H)$ is a $\tau$-$\alpha$ approximate
abstraction of $(M_L, Pr_L)$ if the expectation (in terms of $Pr_L$) of the difference between the states $e$ and $e'$ of the high-level temperature variable $E$ is less or equal to $\alpha$, where $e$ and $e'$ are determined exactly in accordance with the two pathways in Fig.~\ref{fig:approxabstr}, selecting the worst-case  intervention $\vec{W} \gets \vec{w}$ and the best case $\tau_{\U}$.

Analogously to the deterministic case, we have the following straightforward results.
\pro
$M_H$ is a $\tau$-$0$-approximate abstraction of $M_L$ iff  $M_H$ is a
$\tau$-abstraction of $M_L$.  
\epro

\pro
If $M_1$ and $M_2$ are 
similar, then
$M_2$ is a $Id$-$\alpha$-approximate abstraction of $M_1$
iff  $M_2$ is a
$d_{max}$-$\alpha$-approximation of $M_1$.
\epro

\commentout{
  \dfn Given $\tau$-consistent causal models $(M_L, \I_L)$ and  $(M_H,
  \I_H)$, we define an \emph{intervention-distribution ${\Pr}_\I$} to be
  any probability distribution on $\I_L$ conditional on $\I_H$ such that
  for all $\vec{X} \gets \vec{x} \in \I_L$ and $\vec{Y} \gets \vec{y}
  \in \I_H$ the following condition holds: 

  ${\Pr}_\I(\vec{X} \gets \vec{x} | \vec{Y} \gets \vec{y}) > 0$ iff
  $\omega_{\tau}(\vec{X} \gets \vec{x}) = \vec{Y} \gets \vec{y}$.
  \edfn
}%

In Definition \ref{def:dist-prob} we consider the worst-case low-level intervention to define the distance. In many cases, however, 
the whole point of an abstraction is to be able to exclude rare
low-level boundary cases (e.g., when the ideal gas law is taken to
refer only to equilibrium states). Moreover, often the actual
manipulations that we can perform are known to us only at the
high-level, because the low-level implementation of the intervention
is unobservable to us. For example, in setting the room temperature to
$70^{\circ}$F we do not generally consider the instantiation of that
intervention which superheats one corner of the room and freezes the
rest such that the mean kinetic energy works out just right.  
As Spirtes and Scheines \citeyear{SS04}  show, in the absence of any
further information, such ambiguous manipulations can be quite
problematic. 
Fortunately, often our knowledge of the mechanism 
of how  a high-level intervention is implemented does give us significant
probabilistic information.
For example, we might know that the heater has a fan that circulates
the air, most likely resulting in relatively uniform distributions of
the kinetic energies of the particles. 

We capture this information using what we call an intervention distribution.

\dfn Given a surjective map \mbox{$\tau: \R(\V_L) \rightarrow \R(\V_H)$},
%
an \emph{intervention distribution ${\Pr}_\I$} is a distribution on
$\I_L \times\I_H$ such that
${\Pr}_\I(\vec{X} \gets \vec{x} \mid \vec{Y} \gets \vec{y}) > 0$ iff
$\omega_{\tau}(\vec{X} \gets \vec{x}) = \vec{Y} \gets \vec{y}$.
\edfn
We think of $\Pr_\I(\vec{X} \gets \vec{x} \mid \vec{Y} \gets \vec{y})$
as telling us how likely the high-level intervention $\vec{Y}\gets
\vec{y}$ is to have been implemented by the low-level intervention
$\vec{X} \gets \vec{x}$.

\begin{figure}
  \centering
  \includegraphics[width = 6cm]{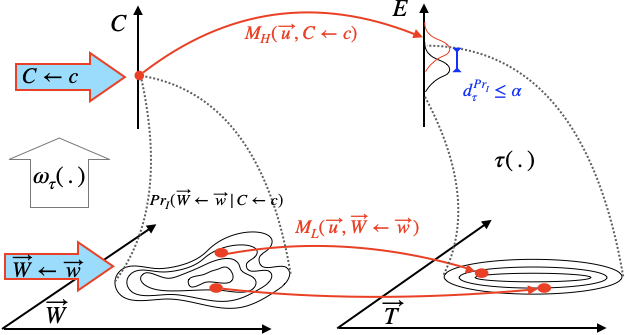}
  \caption{Probabilistic approximate abstraction for the climate
    example: An intervention on the high-level wind variable $C$
    results, given the distribution over contexts $\vec{u}$ (omitted for
    clarity), in the red distribution over the high-level temperature
    variable $E$. The intervention $C\gets c$ can be instantiated in the
    low-level wind map $\vec{W}$ in many ways, according to the
    intervention distribution $Pr_{\I}$. Mapping the (manipulated)
    distribution over $\vec{W}$ using $M_L$ results in a distribution
    over the low-level temperature map $\vec{T}$ that combines
    uncertainty from $Pr_{\I}$ and $Pr_{\U}$. Abstracting this
    distribution using $\tau$ to $E$ results in the black distribution
    over $E$. $M_H$ is a probabilistic $\tau$-$\alpha$-approximate
    abstraction of $M_L$ if the distance between the expectations of
    these distributions is less than $\alpha$.}  
  \label{fig:probapproxabstr} 
\end{figure}

\dfn\label{def:dist-intervention}
Given a surjective map $\tau: \R(\V_L) \rightarrow \R(\V_H)$ and an
intervention-distribution ${\Pr}_\I$, the distance function 
$d_{\tau}^{\Pr_{\I}}$ on
pairs of $\tau$-consistent probabilistic causal models is defined by taking 
$$\begin{array}{ll}
  d_{\tau}^{\Pr_{\I}}(M_L,M_H)= \min_{\{\tau_{\U}: \, 
    \tau_{\U}(\Pr_L)=\Pr_H\}}
  \max_{\vec{Y} \gets \vec{y} \in \I_H} \\
  (
  \sum_{\vec{u}_L \in \R_L(\U_L)} \Pr_L(\vec{u}_L) \\
  \sum_{\vec{X} \gets \vec{x} \in \I_L} {\Pr}_{\I} (\vec{X} \gets
  \vec{x} \mid \vec{Y} \gets \vec{y}) \\ 
  d_{\V_H}(\tau(M_L(\vec{u}_L,\vec{X} \gets
  \vec{x})),M_H(\tau_{\U}(\vec{u}_L),\omega_{\tau}(\vec{X}  \gets  \vec{x})))
  ).
\end{array}$$
\edfn
Intuitively, for each high-level intervention $\vec{Y} \gets \vec{y}$,
we take the expected distance between the two ways of lifting a
low-level intervention to $\vec{Y}\gets \vec{y}$.  In computing the
expectation, there are two sources of uncertainty: the likelihood of a
given context (this is determined by $\Pr_L$) and the likelihood on each
low-level intervention that maps to $\vec{Y} \gets \vec{y}$ (this is
given by $\Pr_{\I}(\cdot \mid \vec{Y} \gets \vec{y})$).
Fig.~\ref{fig:probapproxabstr} illustrates the point for the climate example.

We can also define  $d_{\beta}^{\tau}$ and $d_{\beta}^{\tau,\Pr_{\I}}$
in a manner completely analogous to Definition \ref{def:beta}. We omit
these definitions for reasons of space, but the details should be
clear. 

It is worth comparing our approach to other approaches
to determining the distance between causal models.
The more standard way to compare two causal models is to compare their
causal graphs.  
\commentout{
  (It is typically assumed that the model is
  strongly recursive, so that the ordering $\prec_{\vec{u}}$ is
  independent of the context $\vec{u}$.)
}%
The causal graph
is a directed acyclic graph (dag) that
has nodes labeled by variables; (the node labeled) $X$ is an
ancestor of (the node labeled) $Y$ iff $X \prec Y$.  Dags
have been compared using what is called the \emph{structural Hamming
  distance} (SHD) \cite{AC03},
where the SHD between $G$ and $H$ (which are
assumed to have 
an identical
set of nodes) is the number of pairs of nodes $(i,j)$
on which $G$ and $H$ differ regarding the edge between $i$ and $j$
(either because one of 
one of them
has an edge and the other does not, or the
edges are oriented in different directions).  As Peters and B\"uhlmann
\citeyear{PB15} observe, the SHD misses out on some important
information in causal 
networks. In particular, it does not really
compare the effect of interventions.  They want a notion of distance
that takes this into account, as do we. However,
they take into account the effect of interventions in a way much closer in
spirit to SHD.  Roughly speaking, in our language, given two similar
causal models $M_1$ and $M_2$, they count the
number of pairs 
$(X,Y)$ of endogenous variables such that
intervening on $X$ leads to a different distribution over $Y$ in $M_1$
and $M_2$.  More formally, let $\Pr^{M,X \gets x}_Y$ denote the
marginal of $\Pr^{M,X \gets x}$ on the variable $Y$ in model $M$.
(Since we want to compare probability distributions in two different
models, we add the model to the superscript.)  
The SID between similar models
$M_1$ and $M_2$ is
the number of pairs $(X,Y)$ such that there exists an $x \in
\R(X)$ such that $\Pr^{M_1,X \gets x}_Y \ne \Pr^{M_2,X \gets x}_Y$.

Although 
SID
does compare the predictions that two models make, it
differs from our distance functions in several important respects.
First, it compares just the effect of interventions on single
variables, whereas we allow arbitrary interventions. 
We believe that it is important to consider arbitrary interventions, since
sometimes variables act together, and it takes intervening
on more than one variable to 
 distinguish two models.  Second,
we are interested in how far apart two distributions are, not just the
fact that they are different.  Finally, we want a definition that
applies to models that are not similar, since this is what we need for
approximate abstraction.

%
\section{COMPOSING ABSTRACTION AND APPROXIMATION}\label{sec:compose}

It can be useful to understand an approximate abstraction as the
result of composing an approximation and an abstraction, in some
order.  For example, we can explain the ideal gas law in terms of
thinking of frictionless elastic collisions between particles (this is
an approximation to the truth) and then abstracting by replacing the
kinetic energy of the particles by their temperature (a measure of
average kinetic energy).  
Here we examine the extent to
which an approximate abstraction can be viewed this way.  We start
with two easy results showing that if we compose an approximation and
an abstraction in some order, then we do get an approximate abstraction.

\pro\label{pro:comp1}
If $(M_H, \I_H)$ is a $\tau$-abstraction of $(M_L, \I_L)$ and $(M_H',
\I_H)$ is a $d_{max}$-$\alpha$-approximation of $(M_H, \I_H)$ then $(M_H',
\I_H)$ is a $\tau$-$\alpha$ approximate abstraction of $(M_L, \I_L)$.%
\shortv{
  \footnote{Proofs of all technical results
    can be found in the 
        full paper \cite{BEH19a}.} 
}
\fullv{
  \footnote{Proofs of all technical results
    from here onwards
    can be found in the 
    appendix.}
}
\epro

In Proposition~\ref{pro:comp1} we considered an abstraction followed
by an approximation.  Things change if we do things in the opposite
order; that is, if 
we consider an approximation followed by an abstraction.  For suppose that $M_L'$ is a
$d_{max}$-$\alpha$ approximation of $M_L$ and $M_H$ is a $\tau$-abstraction of $M_L$. 
Now it is not in general the case that
$M_H$ is a $d_{\tau}$-$\alpha$-approximate abstraction of $M_L'$.  The
problem is that when assessing how good an abstraction $M_H$ is of
$M_L'$, we are comparing two high-level states (using $d_{\V_H}$).  But when
comparing $M_L$ to $M_L'$, we use $d_{\V_L}$.  In general,
$d_{\V_L}(\vec{v}_L,\vec{v}_L')$ and $d_{\V_H}(\tau(\vec{v}_L), \tau(\vec{v}_L'))$ may be unrelated.
%

\pro\label{pro:comp2}
If $(M_H, \I_H)$ is a $\tau$-abstraction of $(M_L, \I_L)$ and $(M_L,\I_L)$
is a $d_{max}$-$\alpha$ approximation of $(M_L', \I_L)$, 
then $(M_H, \I_H)$ is a $\tau$-$k\alpha$ approximate
abstraction of $(M_L', \I_L')$, where $k$ is
$$\begin{array}{ll}  
  \max_{\vec{X} \gets \vec{x} \in \I_L, \vec{u}_L \in \R_L(\U_L)} \\
  \frac{d_{\V_H}(\tau(M_L'(\vec{u}_L,\vec{X} \gets
    \vec{x})),\tau(M_L(\vec{u}_L,\vec{X}  \gets \vec{x})))}{
    d_{\V_L}(M_L'(\vec{u}_L,\vec{X} \gets
    \vec{x}),M_L(\vec{u}_L,\vec{X}  \gets \vec{x}))}.
\end{array}$$
\epro

\commentout{
  We are typically more interested in viewing an approximate abstraction
  as the abstraction of an approximation or the approximation of an
  abstraction.  As we show by example below, this may not be possible.
  As the following result shows, we can determine whether it is
  possible in nondeterministic polynomial time.
  [[I STRONGLY SUSPECT THAT IT'S NP COMPLETE, BUT I HAVEN'T THOUGHT THIS
      THROUGH 
    }%
    While composing an approximation and an abstraction gives us an
    approximate abstraction, 
    \shortv{the full paper \citeyear{BEH19a} contains two examples 
    that show that we
    cannot in general decompose an approximate 
    abstraction into an abstraction composed with an approximation or 
    an approximation composed with an abstraction.
    \commentout{
    Theorem~\ref{th:const} below shows that if we restrict ourselves to
    the constructive case then
    we can do the former. 
    }%
      }
    \fullv{
    the following two examples show that we
    cannot in general decompose an approximate 
    abstraction into an abstraction composed with an approximation or 
    an approximation composed with an abstraction.
    Theorem~\ref{th:const} below shows that if we restrict ourselves to
    the constructive case then
    we can do the former. 
    }
    \fullv{
    However,
    Example~\ref{xam: no construction} shows that, even if we restrict to
    the constructive case, we cannot do the latter.
    }

    
    \fullv{
    \xam\label{xam:new3}
    $M_L$ has one exogenous variable $U$ and endogenous variables $A$, $B$,
    and $C$.  $M_H$ also has one exogenous variable $U$ and endogenous
    variables $D$ and $E$.  All the endogenous variables have range
    $\{0,1\}$; $U$ has range $\{0\}$.
    Let $\tau:\R(\V_L) \rightarrow \R(\V_H)$ be
    $\tau(a,b,c)=(a\oplus b,b \oplus c)$, where $\oplus$ denotes addition
    mod 2.
    The equations for both
    $M_L$ and $M_H$ take the form $V=U$ for all variables $V$. $\I_L$
    consists of the empty intervention,  and interventions $B \gets 1$, $B
    \gets 0$, $A \gets 1$, and $A \gets 0$. $\I_H$ consists of only the empty
    intervention. Taking as $d_{\V_H}$ the Euclidean distance, we leave it
    to the reader to verify that $M_H$ is a $\tau$-$\sqrt{2}$ approximate
    abstraction of $M_L$. However, there does not exist any $M_L'$ that is
    similar to $M_L$ such that $M_H$ is a $\tau$-abstraction of $M_L'$. To
    see why, note that the only state in $M_H$ that arises from applying
    an intervention is $(0,0)$.
    Therefore, the only states in $M_L$ that can arise from interventions are
    ones where $a \oplus b=0$ and $b \oplus c=0$. This means that under
    the intervention 
    $B \gets 0$, we must 
    have $A=0$, and under $B \gets 1$, we must have $A=1$. Therefore $B
    \prec A$.
    It also means that under $A \gets 1$ we must have
    $B=1$ and under $A \gets 0$ we must have $B=0$, so that 
    $A \prec
    B$.  Thus, $M_H$ cannot be recursive.
    \exam

    \xam\label{xam: no construction}
    Let $M_L$ be the model where $\U_L = \{U\}$, $\V_L = \{X_1,
    X_2, X_3\}$, $\R(X_i) = \{0,1\}$ for $i = 1, 2, 3$, the equations
    are $X_1 = U$, $X_2 = X_1$, and $X_3 = X_2$, and $\I_L$ consists of
    all interventions such that $X_1$ is intervened on iff $X_3$ is
    intervened on.   Let $M_H$ be the model
    where $\U_H = \U_L = \{U\}$, $\V_H = \{Y_1,Y_2\}$,
    $\R(Y_1) = \{0,1\} \times \{0,1\}$, $\R(Y_2) = \{0,1\}$, the equations 
    are $Y_1 = (U,U)$, $Y_2 = \mathit{proj}_1(Y_1)$, the first component
    of $Y_1$, and $\I_H$ consists of all interventions.
    Let $P= \{\{X_1,X_3\}, \{X_2\}\}$, and 
    $\tau$ is such that $\tau(x_1,x_2,x_3) = ((x_1,x_3),x_2)$.
    Note that $\tau$ is constructive,
    that $M_H$ and $M_L$ are $\tau$-consistent, and that
    $\I_L=\I_L^{\tau}$. Therefore $M_H$ is a constructive $\tau$-$\alpha$
    approximate abstraction of $M_L$ for some $\alpha$ (the value of
    $\alpha$ depends on the choice of metric $d_{\V_H}$). 
    It is easy to see that $\omega_\tau(X_1 \gets i, X_3 \gets j) = (Y_1
    \gets (i,j))$ and $\omega_\tau(X_2 \gets k) = (Y_2 \gets k)$.  
    Note that $M_L(0,(X_1 \gets 1, X_3 \gets 0)) = (1,1,0)$,
    $M_L(0,(X_1 \gets 0, X_3 \gets 1)) = (0,0,1)$, 
    $M_L(0,X_2 \gets 0) = (0,0,0)$, and $M_L(0,X_2 \gets 1) = (0,1,1)$.
    For a causal model $M_H'$ that is similar to $M_H$ to be a $\tau$-abstraction of
    $M_L$, there must be some 
    surjection $\tau_\U$ such that $M_H(\tau_\U(0), Y_1 \gets (1,0)) = ((1,0),1)$,
    $M_H(\tau_\U(0), Y_1 \gets (0,1)) = ((0,1),0)$,
    $M_H(\tau_\U(0), Y_2 \gets 0) = ((0,0),0)$, and
    $M_H(\tau_\U(0), Y_2 \gets 1) = ((0,1),1)$.  It follows that
    $Y_1 \prec_{\tau_\U(0)} Y_2$ and $Y_2 \prec_{\tau_\U(0)} Y_1$.
    Thus, there is no (recursive) 
    model 
    $M_H'$ that is a
    $\tau$-abstraction of $M_L$.
    \exam
}

    As the following result shows, we can test whether an approximate abstraction can be viewed as the result of composing an abstraction and an approximation.

    \thm\label{pro:comp3} If $(M_H, \I_H)$ is a
    $\tau$-$\alpha$-approximate abstraction of 
    $(M_L, \I_L)$  
    then the problem of determining
    whether there exists a model $(M_H',\I_H)$ (resp., $(M_L',\I_L)$) 
    that is similar to $(M_H, \I_H)$
    and 
    is an abstraction of $(M_L,\I_L)$ (resp., $(M_L,\I_L)$) 
    is in nondeterministic
    time polynomial in $|\I_L| \times |\R_L(\U_L)|$.  
    \ethm

    \commentout{
      [SOME THOUGHTS: Note that this result leaves open that possibility
        that you can do get a better approximation by going diagonally
        rather than by going up and across.  That is, $M_H$ there may be a
        $\tau$-$\alpha$-approximate abstraction  of $M_L'$ and there may
        exist $M_H'$ such that $M_H'$ is an exact abstraction of $M_L'$ but 
        there may not exist an exact abstraction $M_H'$ such that $M_H'$ is
        an $\alpha$-approximation of $M_L$. Note also if the models are
        relatively small, we can generate all the abstractions and check
        which is the best approximation.  I'm not sure that we can  do
        better than that.]
    }


    There is one important special case where we are guaranteed to be able
    to find an appropriate intermediate low-level model, and to do so in
    polynomial time: if the mapping $\tau$ is constructive.

    %
    \thm\label{th:const} If $\tau: \R_L(\V_L) \rightarrow \R_H(\V_H)$ is
    constructive,  
    the causal models  $(M_L, \I_L)$ and 
    $(M_H, \I_H)$ are
    $\tau$-consistent, and 
    $(M_H, \I_H)$ is a
    $\tau$-$\alpha$-approximate abstraction of $(M_L, \I_L)$, then we can find
    a  model $(M'_L, \I_L)$ that 
    is similar to $(M_L, \I_L)$,
    and such that  $(M_H,
    \I_H)$ is a $\tau$-abstraction of $(M'_L, \I_L)$
    in time polynomial in $|\I_L| \times |\R_L'(\U_L)|$.  
    \ethm

    \shortv{The full paper contains an example showing that 
    $\tau$
    being constructive does not similarly guarantee the 
     existence of $M_H'$ in the first half of Theorem~\ref{pro:comp3}.    }
    \fullv{
    As noted earlier, Example~\ref{xam: no construction} shows that $\tau$
    being constructive does not similarly guarantee the 
     existence of $M_H'$ in the first half of Theorem~\ref{pro:comp3}.
    }

    \commentout{
      \subsection{Algorithms}

      Formulate optimization problem: given $(M_L, \I_L)$, $\tau: \R_L(\V_L)
      \rightarrow \R(\V_H)$, and a metric $d_H$, find the smallest $\alpha$
      such that there exists a $(M_H, \I_H)$ which is a
      $\tau$-$\alpha$-approximate abstraction of $(M_L, \I_L)$. 

      The algorithm will construct the required $(M_H, \I_H)$. 
    }


    \section{DISCUSSION AND CONCLUSIONS}
    By defining notions of abstraction, approximation, and approximate
    abstraction, we have presented a framework that relates causal models
    that describe the same system at  
    (possibly)
    different levels of granularity. While coarser models offer a degree
    of simplification by omitting details, they also in general entail a
    loss in accuracy with respect to the fundamental description. Our
    framework shows how to quantify this loss in accuracy by defining a
    distance metric that captures the degree to which a more abstract
    causal model approximates a more detailed causal model. High- and
    low-level causal models of the same system can vary on almost any
    dimension. They need not share the same equations, the same
    variables, or the same interventions. They may involve entirely distinct state
    spaces. Inevitably, then, there is some degree of choice  
    as to
    what one deems relevant to the approximation.

    Starting with
    deterministic causal models, we provided a general method
    for  quantifying the ``goodness'' of an approximate abstraction.
    As an interesting special case, our approach allows for the comparison of
    causal models that operate at the same level of detail. 
    We then extended to probabilistic causal models, and considered several
    different choices  
    for quantifying the distance between models.
    Finally, we considered the extent to which we could decompose an
    approximate abstraction into an abstraction and an approximation. 

    Given the ubiquitous use of causal models in the social and natural
    sciences that are known not to capture all the causally relevant
    details, the framework we presented offers a principled way to assess
    the trade-off between abstraction and accuracy. 

    
    \fullv{
      \appendix
      \section*{Appendix: Proofs}

      In this appendix, we prove all the results not proved in the main text.
      We repeat the statements of the results for the reader's convenience

      \newenvironment{oldthm}[1]{\par\noindent{\bf Theorem #1:} \em \noindent}{\par}
      \newenvironment{oldlem}[1]{\par\noindent{\bf Lemma #1:}
        \em \noindent}{\par}
      \newenvironment{oldpro}[1]{\par\noindent{\bf Proposition #1:}
        \em \noindent}{\par}
      \newcommand{\othm}[1]{\begin{oldthm}{\ref{#1}}}
      \newcommand{\eothm}{\end{oldthm} \medskip}
      \newcommand{\olem}[1]{\begin{oldlem}{\ref{#1}}}
      \newcommand{\eolem}{\end{oldlem} \medskip}
      \newcommand{\opro}[1]{\begin{oldpro}{\ref{#1}}}
      \newcommand{\eopro}{\end{oldpro} \medskip}

      \medskip
      \opro{pro:comp1}
      If $(M_H, \I_H)$ is a $\tau$-abstraction of $(M_L, \I_L)$ and $(M_H',
      \I_H)$ is a $d_{max}$-$\alpha$-approximation of $(M_H, \I_H)$ then $(M_H',
      \I_H)$ is a $\tau$-$\alpha$ approximate abstraction of $(M_L, \I_L)$.
      \eopro

      \prf  Fix $\vec{X} \gets \vec{x} \in \I_L$ and $\vec{u}_L \in \R_L(\U_L)$.
      Since $(M_H,\I_H)$ is a $\tau$-abstraction of $(M_L,\I_L)$,
      there is a mapping $\tau_{\U}: \U_L \rightarrow \U_H$ such that 
      $$\tau(M_L(\vec{u}_L,\vec{X}  \gets \vec{x})) =
      M_H(\tau_{\U}(\vec{u}_L),\omega_{\tau}(\vec{X} \gets \vec{x})).$$ 
      Since $(M_H',\I_H)$ is a $d_{max}$-$\alpha$ approximation of
      $(M_H,\I_H)$, we must 
      have 
      $$\begin{array}{ll}  
        &d_{\V_H}(M_H'(\tau_{\U}(\vec{u}_L),\omega_{\tau}(\vec{X} \gets
        \vec{x}))),\tau(M_L(\vec{u}_L,\vec{X}  \gets \vec{x})))\\
        = &d_{\V_H}(M_H'(\tau_{\U}(\vec{u}_L),\omega_{\tau}(\vec{X} \gets
        \vec{x}))),
        \\ & \ \ M_H(\tau_{\U}(\vec{u}_L),\omega_{\tau}(\vec{X} \gets
        \vec{x}))) \\
        \le &\alpha.\end{array}$$
      Thus, $(M_{H'},\I_H)$ is a $\tau$-$\alpha$ approximate abstraction of
      $(M_L, \I_L)$.
      \eprf

      \opro{pro:comp2}
      If $(M_H, \I_H)$ is a $\tau$-abstraction of $(M_L, \I_L)$ and $(M_L,\I_L)$
      is a $d_{max}$-$\alpha$ approximation of $(M_L', \I_L)$, 
      then $(M_H, \I_H)$ is a $\tau$-$k\alpha$ approximate
      abstraction of $(M_L', \I_L')$, where $k$ is
      $$\begin{array}{ll}  
        \max_{\vec{X} \gets \vec{x} \in \I_L, \vec{u}_L \in \R_L(\U_L)} \\
        \frac{d_{\V_H}(\tau(M_L'(\vec{u}_L,\vec{X} \gets
          \vec{x})),\tau(M_L(\vec{u}_L,\vec{X}  \gets \vec{x})))}{
          d_{\V_L}(M_L'(\vec{u}_L,\vec{X} \gets
          \vec{x}),M_L(\vec{u}_L,\vec{X}  \gets \vec{x}))}.
      \end{array}$$
      \eopro

      \prf  Fix $\vec{X} \gets \vec{x} \in \I_L$ and $\vec{u}_L \in \R_L(\U_L)$.
      Since $(M_H,\I_H)$ is a $\tau$-abstraction of $(M_L,\I_L)$,
      there is a mapping $\tau_{\U}: \U_L \rightarrow \U_H$ such that 
      $$\tau(M_L(\vec{u}_L,\vec{X}  \gets \vec{x})) =
      M_H(\tau_{\U}(\vec{u}_L),\omega_{\tau}(\vec{X} \gets \vec{x})).$$ 
      Since $M_L'$ is a $d_{max}$-$\alpha$ approximation of $M_L$, we 
      have that
      $d_{\V_L}(M_L'(\vec{u}_L,\vec{X} \gets
      \vec{x}),M_L(\vec{u}_L,\vec{X}  \gets \vec{x})) \le \alpha$.
      By the definition of $k$, we have that 
      $d_{\V_H}(\tau(M_L'(\vec{u}_L,\vec{X} \gets
      \vec{x})),\tau(M_L(\vec{u}_L,\vec{X}  \gets \vec{x})) \le k\alpha$.
      Thus, 
      $d_{\V_H}(M_H(\tau_{\U}(\vec{u}_L),\omega_{\tau}(\vec{X} \gets
      \vec{x})),\tau(M_L(\vec{u}_L,\vec{X}  \gets \vec{x})) \le k\alpha$. 
      It follows that $(M_H,\I_H)$ is a $\tau$-$k\alpha$ approximate
      abstraction of $(M_L',\I_L')$.
      \eprf

      To make our results more general, we use the more general
      interpretation of recursiveness as it appears in \cite{Hal48}.
      Concretely, this means that  the  partial order $\preceq$ on the
      endogenous variables may depend on the context. We write
      $\preceq_{\vec{u}}$ for the partial order that exists for context
      $\vec{u}$. (See footnote \ref{note:recursive}.)

      \medskip
      \othm{pro:comp3} If $(M_H, \I_H)$ is a
      $\tau$-$\alpha$-approximate abstraction of 
      $(M_L, \I_L)$ and $\U_L$ and $\I_L$
      are finite, 
      then the problem of determining
      whether there exists a model $(M_H',\I_H)$ (resp., $(M_L',\I_L)$) 
      that is similar to $(M_H, \I_H)$
      and 
      is an abstraction of $(M_L,\I_L)$ (resp., $(M_L,\I_L)$ 
      is in nondeterministic
      time polynomial in $|\I_L| \times |\R_L'(\U_L)|$.  
      \eothm
      \medskip

      \prf
      We start with the problem of determining $M_H'$.
      Since $\U_H$ and $\U_L$ are finite, there are only finitely many
      possible surjections from $\U_L$ to $\U_H$.  A surjection
      $\tau_{\U}$ is \emph{potentially high-level compatible with $\tau$} if 
      %
      \begin{enumerate}
      \item[PC1.] For all $\vec{X} \gets \vec{x}, \ \vec{X}' \gets \vec{x}' \in
        \I_L$ and all $\vec{u},\, \vec{u}' \in \R_L(\U_L)$,
        if $\omega_\tau(\vec{X} \gets \vec{x}) = \omega_\tau(\vec{X}'
        \gets \vec{x}')$ and $\tau_{\U}(\vec{u}) = \tau_{\U}(\vec{u}')$, then 
        $\tau(M_L(\vec{u}_L,\vec{X} \gets \vec{x})) =
        \tau(M_L(\vec{u}_L',\vec{X}' \gets \vec{x}'))$ 
      \item[PC2.] For all $\vec{u}_H \in \R_H(\U_H)$, there exists a partial
        order $\prec_{\vec{u}_H}$ on the variables in $\V_H$ such that
        for all $\vec{u}_L \in \R_L(\U_L)$ with $\vec{u}_H =
        \tau_{\U}(\vec{u}_L)$, all pairs of 
        interventions  $\vec{X} \gets \vec{x}$ and $\vec{X}' \gets \vec{x}'$
        in $\I_L$, and all variables $Y \in \V_H$ whose  value differs 
        in $\tau(M_L(\vec{u}_L,\vec{X} \gets \vec{x}))$
        and $\tau(M_L(\vec{u}_L,\vec{X}'
        \gets \vec{x}'))$, there exists a variable $Z$ such that
        $Z \preceq_{\vec{u}_H} Y$ (i.e., $Z \prec_{\vec{u}_H} Y$ or $Z=Y$) and
        different values in $\omega_\tau(\vec{X} \gets \vec{x})$ and
        $\omega_{\tau}(\vec{X}' \gets \vec{x}')$ or $Z$ is assigned a value in one
        of $\omega_\tau(\vec{X} \gets \vec{x})$ and
        $\omega_\tau(\vec{X}' \gets \vec{x}')$ and not in the other.   
      \end{enumerate}

      We claim that $\tau_{\U}$ is potentially high-level compatible with $\tau$
      iff there exists a (recursive) causal model
      $(M_H',\I_H)$ such that, for all contexts $\vec{u}_L \in \R_L(\U_L)$
      and interventions $\vec{X} \gets \vec{x} \in \I_L$, we have 
      \begin{equation}\label{eq1}
        \tau(M_L(\vec{u}_L,\vec{X}  \gets \vec{x})) =
        M_H'(\tau_{\U}(\vec{u}_L),\omega_{\tau}(\vec{X} \gets \vec{x})).
      \end{equation}

      It is easy to see that if PC1 or PC2 do not hold for $\tau_{\U}$,
      then there can be no 
      (recursive) causal model $(M_H',\I_H)$ satisfying (\ref{eq1}).  For
      the converse, 
      to build the model $M_H'$, we have to specify the equations for
      each variable in such a way that $\prec_{\vec{u}_H}$ really is the
      partial order showing the dependence on variables in context $\vec{u}_H$.
      Say that a high-level state $\vec{v}_H \in \R_H(\V_H)$ is
      \emph{constructible for $\vec{u}_H$ and $Y \in \V_H$} if
      $\vec{v}_H = \tau(M_L(\vec{u}_L,\vec{X}  \gets \vec{x}))$ for some
      intervention $\vec{X} \gets \vec{x} \in \I_L$ and context $\vec{u}_L
      \in \R_L(\U_L)$ such that $\tau_\U(\vec{u}_L) = \vec{u}_H$ and
      $\omega_\tau(\vec{X} \gets \vec{x})$ does not include an intervention
      on $Y$.  Each high-level state 
      $\vec{v}_H$  constructible for $\vec{u}_H$ 
      and $Y \in \V_H$ determines one output of $F'_Y$ 
      in context $\vec{u}_H$ in the obvious way.  
      Specifically, if $F'_Y$ gets as arguments $\vec{u}_H$ and the 
      values of the variables other than $Y$ in $\vec{v}_H$, then it must
      output the value of $Y$ in $\vec{v}_H$.  Note that it follows from PC2
      that for the values of $F'_Y$ so defined, if two inputs to $F'_Y$ agree
      on the values of all variables $Z$ such that $Z \prec_{\vec{u}_H} Y$,
      they will agree on the value of $Y$.  We want to extend all the
      equations $F'_Y$ such that this continues to be true.  This is
      straightforward.

      Fix $\vec{u}_H$.  We define $F_Y'$ when the context is $\vec{u}_H$ for
      all variables $Y \in \V_H$ as follows.
      If $Y$ has no predecessors
      in the $\prec_{\vec{u}_H}$ order, then it gets the same value in all the 
      constructible states for 
      context $\vec{u}_H$.  We extend $F_Y'$ so that $Y$ gets that
      value no matter what the values of the endogenous variables other than
      $Y$ are in context
      $\vec{u}_H$.
      Similarly, if $Z_1, \ldots, Z_k$ are the
      variables that precede $Y$ in the $\prec_{\vec{u}_H}$ order, 
      and $(Z_1,\ldots, Z_k) = (z_1, \ldots, z_k)$ appears in some constructible 
      state for $\vec{u}_H$ then, by PC2, $Y$ has the same value in all constructible
      states for $\vec{u}_H$ where $(Z_1,\ldots, Z_k) = (z_1, \ldots, z_k)$.  
      We extend $F_Y'$ so that $Y$ has that value for all inputs  
      where $(Z_1,\ldots, Z_k) = (z_1, \ldots, z_k)$ and the context is $\vec{u}_H$.
      If there is no constructible state where $(Z_1,\ldots, Z_k) = (z_1,
      \ldots, z_k)$, then we just pick a fixed value  $y \in \R_H(Y)$ and
      take $F_Y'$ to be $y$ for all inputs where
      $(Z_1,\ldots, Z_k) = (z_1, \ldots, z_k)$ and the context is
      $\vec{u}_H$.  It is clear by construction that this definition has the
      desired properties.

      We conclude this part of the proof by observing that checking that PC1
      holds can be 
      done in time polynomial in $|\I_L| \times |\R_L(\U_L)|$, and for a
      fixed ordering $\prec_{\tau_{\U}}$, PC2 can be checked in time
      polynomial in $|\I_L| \times |\R_L(\U_L)|$.  Thus, we can determine whether there
      exists a model $(M_H',\I_H)$ that is an abstraction of $(M_L,
      \I_L)$ and differs from $(M_H,\I_H)$ only in the equations by guessing
      $\tau_{\U}$ and 
      a collection of partial orders $\prec_{\vec{u}_H}$, one for each
      high-level context, and confirming that PC1 and PC2 hold.   Thus, this
      can be done in nondeterministic time polynomial in $|\I_L| \times
      |\R_L(\U_L)|$.  

      The algorithm for determining $(M_L',\I_L)$ is very similar in spirit.
      A surjection $\tau_{\U}: \R_L(\V_L) \rightarrow \R_H(\V_H)$ and a
      function $f: \R(\U_L) \times \I_L \rightarrow \R(\V_L)$
      are \emph{potentially low-level compatible} with $\tau$ if they satisfy:
      \begin{enumerate}
      \item[PC1$'$.] For all $\vec{X} \gets \vec{x}, \ \vec{X}' \gets \vec{x}' \in
        \I_L$ and all $\vec{u},\, \vec{u}' \in \R_L(\U_L)$,
        if $\omega_\tau(\vec{X} \gets \vec{x}) = \omega_\tau(\vec{X}'
        \gets \vec{x}')$ and $\tau_{\U}(\vec{u}) = \tau_{\U}(\vec{u}')$, then 
        $\tau(f(\vec{u}_L,\vec{X} \gets \vec{x})) =
        \tau(f(\vec{u}_L',\vec{X}' \gets \vec{x}'))$ 
      \item[PC2$'$.] For all $\vec{u}_L \in \R_L(\U_L)$, there exists a partial
        order $\prec_{\vec{u}_L}$ on the variables in $\V_L$ such that
        for all interventions  $\vec{X} \gets \vec{x}$ and $\vec{X}' \gets \vec{x}'$
        in $\I_L$, and all variables $Y \in \V_L$ whose  value differs in 
        $f(\vec{u}_L,\vec{X} \gets \vec{x})$ and $f(\vec{u}_L,\vec{X}'
        \preceq_{\vec{u}_L} Y$ and either $Z$ gets 
        different values in $\vec{X} \gets \vec{x}$ and
        $\vec{X}' \gets \vec{x}'$ or $Z$ is assigned a value in one
        of $\vec{X} \gets \vec{x}$ and
        $\vec{X}' \gets \vec{x}'$ and not in the other.   
      \end{enumerate}
      Essentially, $f(\vec{u}_L,\vec{X} \gets \vec{x})$ is playing the same
      role in PC2$'$ as $M_L(\vec{u}_L,\vec{X} \gets \vec{x})$ played in
      PC2.  

      We now claim that $\tau_{\U}: \R_L'(\V_L') \rightarrow \R_H(\V_H)$ and 
      $f: \R(\U_L') \times \I_L' \rightarrow \R(\V_L')$
      are \emph{potentially low-level compatible} with $\tau$
      iff there exists a (recursive) causal model
      $(M_L',\I_L)$ such that 
      for all contexts $\vec{u}_L \in \R_L(\U_L)$
      and interventions $\vec{X} \gets \vec{x} \in \I_L$, we have 
      $M_L'(\vec{u}_L,\vec{X}  \gets \vec{x})) =
      f(\vec{u}_L,\vec{X}  \gets \vec{x}))$  and
      $$
      \tau(M_L'(\vec{u}_L,\vec{X}  \gets \vec{x})) =
      M_H(\tau_{\U}(\vec{u}_L),\omega_{\tau}(\vec{X} \gets \vec{x})).
      $$
      The argument is almost identical to that given above for the first
      part, so we omit it here.

      Thus, to determine whether there exists an appropriate model
      $(M_L',\I_L)$, we simply need to guess $f$, $\tau_\U$, and
      $\prec_{\vec{u}_L}$ for all $\vec{u}_L \in \R_L(\U_L)$ and verify
      that PC1$'$ and PC2$'$ hold.
      \eprf

      \medskip
      \othm{th:const} If $\tau: \R_L(\V_L) \rightarrow \R_H(\V_H)$ is constructive, 
      the causal models  $(M_L, \I_L)$ and 
      $(M_H, \I_H)$ are
      $\tau$-consistent, and 
    $(M_H, \I_H)$ is a
      $\tau$-$\alpha$-approximate abstraction of $(M_L, \I_L)$, then we can find
      a  model $(M'_L, \I_L)$ that 
      is similar to $(M_L, \I_L)$,
      and such that $(M_H,
      \I_H)$ is a $\tau$-abstraction of $(M'_L, \I_L)$
      in time polynomial in $|\I_L| \times |\R_L'(\U_L)|$.  
      \eothm

      \prf Whereas in the proof of the second half in
      Theorem~\ref{pro:comp3} we had to guess $\tau_\U$ and $f$, here we
      can construct them efficiently.  Indeed, we can take $\tau_\U$ to be
      an arbitrary surjection from $\U_L$ to $\U_H$.

      To define $f$, suppose that $\V_H
      = \{Y_1, \ldots, Y_n\}$ and $P_\tau = \{\vec{Z}_1, \ldots,
      \vec{Z}_{n+1}\}$ is the partition that makes $\tau$ constructive (as
      in Definition~\ref{dfn:constructive}).  

      In constructing $f(\vec{u}_L,\vec{X} \gets \vec{x})$, we split the
      intervention $\vec{X} \gets \vec{x}$ into two parts: an intervention
      on variables in $\vec{Z}_1 \cup \ldots \cup \vec{Z}_n$ and an 
      intervention on variables in $\vec{Z}_{n+1}$.  How $f$ works in the
      former case is determined by translating the intervention to $M_H$.
      Interventions on variables in $\vec{Z}_{n+1}$ are 
      treated specially.  An intervention $V \gets v$ on a variable $V \in
      \vec{Z}_{n+1}$ just sets $V$ to $v$, and does not affect any other
      variables.    In more detail, we proceed as follows.

      Note that $\tau$ is also a surjection from $\R_L(\vec{Z}_1 \cup
      \ldots \cup\vec{Z}_n)$ to $\R_H(\V_H)$, since the variables in
      $\vec{Z}_{n+1}$ are ignored by $\tau$.
      Thus, there is a right inverse $\tau^{-1}: 
      \R_H(\V_H) \rightarrow \R_L(\vec{Z}_1 \cup \ldots cup \vec{Z}_n)$ (so
      that $\tau \circ \tau^{-1}$ is 
      the identity function on $\R_H(\V_H)$).  There are, in
      general, many such left inverses, but given $\tau$, we can find a left
      inverse in time polynomial $|\R(\U_L)|$.

      Fix  a setting $\vec{z}_{n+1}^*$ for the variables in $\vec{Z}_{n+1}$.
      Given an intervention 
      $\vec{X} \gets \vec{x}$, let $\vec{X}^\dagger$  be the subset of
      variables in $\vec{X}$ that are in $\vec{Z}_1 \cup \ldots \cup
      \vec{Z}_n$, and let $\vec{X}^{\dagger\dagger}$ be the subset of variables in
      $\vec{X}$ that are in $\vec{Z}_{n+1}$.  Let $\vec{x}^\dagger$ and $\vec{x}^{\dagger\dagger}$
      be the restrictions of $\vec{x}$ to $\vec{X}^{\dagger}$ and $\vec{X}^{\dagger\dagger}$,
      respectively.  Define $$\begin{array}{ll}
        f(\vec{u}_L, \vec{X}\gets\vec{x}) =\\
        (\tau^{-1}(M_H(\tau_\U(\vec{u}_L), \omega_T(\vec{X}^{\dagger}\gets
        \vec{x}^{\dagger}))), \vec{z}_{n+1}^*[\vec{X}^{\dagger\dagger} =
          \vec{x}^{\dagger\dagger}]),\end{array}$$ where  
      $\vec{z}_{n+1}^*[\vec{X}^{\dagger\dagger} = \vec{x}^{\dagger\dagger})]$ is the tuple 
      that results from  $\vec{z}_{n+1}^*$ by setting the values of the variables
      in $\vec{X}^{\dagger\dagger}$ to $\vec{x}^{\dagger\dagger}$.

      We first check PC1$'$. Suppose that $\vec{X} \gets \vec{x}, \ \vec{X}'
      \gets \vec{x}' \in \I_L$, $\vec{u}_L,\, \vec{u}_L' \in \R_L(\U_L)$,
      $\omega_\tau(\vec{X} \gets \vec{x}) = \omega_\tau(\vec{X}'
      \gets \vec{x}')$, and $\tau_{\U}(\vec{u}_L) =
      \tau_{\U}(\vec{u}_L')$.
      Then, using the same notation as above and just writing ``$\ldots$'' for
      the component of the state describing the values of the variables in
      $\vec{Z}_{n+1}$, since these values are ignored by $\tau$, 
      we have that 
      $$\begin{array}{ll}
        &\tau(f(\vec{u}_L,\vec{X} \gets \vec{x})) \\
        =
        &\tau(\tau^{-1}(M_H(\tau_{\U}(\vec{u}_L),\omega_{\tau}(\vec{X}^{\dagger}
        \gets \vec{x}^{\dagger}))),\ldots)\\
        = &\tau(\tau^{-1}(M_H(\tau_{\U}(\vec{u}'_L),\omega_{\tau}((\vec{X}')^{\dagger}
        \gets (\vec{x}')^{\dagger}))),\ldots)
        \\ =&\tau(f(\vec{u}'_L,\vec{X}' \gets \vec{x}')).
      \end{array}$$

      To see that PC2$'$ holds, fix $\vec{u}_L \in \R_L(\U_L)$.
      Suppose that $\V_H
      = \{Y_1, \ldots, Y_n\}$ and $P_\tau = \{\vec{Z}_1, \ldots,
      \vec{Z}_{n+1}\}$ is the partition that makes $\tau$ constructive (as
      in Definition~\ref{dfn:constructive}).  
      Since $M_H$
      is a recursive causal model, there exists a partial 
      order $\prec_{\tau_{\U}(\vec{u}_L)}$ on the variables in $\V_H$ such
      that if $Y_i$ depends on $Y_j$ in 
      context $\tau_{\U}(\vec{u}_L)$, then $Y_j \prec_{\tau_{\U}(\vec{u}_L)} Y_i$. 
      Define $\prec_{\vec{u}_L}$ so that $X \prec_{\vec{u}_L} X'$ iff
      for some $i$ and $j$, $X \in Z_j$, $X' \in Z_i$, and
      $Y_j \prec_{\tau_{\U}(\vec{u}_L)} Y_i$.
      (Note that this means that the variables in $\vec{Z}_{n+1}$ are
      incomparable to all the rest.)

      We now show that this choice of $\prec_{\vec{u}_L}$ satisfies PC2$'$.
      %
      Suppose that $\vec{X} \gets \vec{x}$ and $\vec{X}' \gets \vec{x}'$
      in $\I_L$, and the value of 
      $V \in \V_L$ differs in 
      $f(\vec{u}_L,\vec{X} \gets \vec{x})$
      and $f(\vec{u}_L,\vec{X}' \gets \vec{x}')$.

      If $V \in \vec{Z}_{n+1}$, then it must be the case that $V$ is in
      $\vec{X}$ or $\vec{X}'$.  
      On the other hand, if $V \in \vec{Z}_i$ for some $i \le n$, then
      the value of $Y_i$ differs in 
      $M_H(\tau_{\U}(\vec{u}_L),\omega_{\tau}(\vec{X} \gets \vec{x}))$ and
      $M_H(\tau_{\U}(\vec{u}_L),\omega_{\tau}(\vec{X}' \gets \vec{x}'))$.  
      %
      Therefore, there exists a variable $Y_j \in \V_H$ such that $Y_j
      \prec_{\tau_{\U}(\vec{u}_L)} Y_i$ and either $Y_j$ gets  
      different values in $\omega_{\tau}(\vec{X} \gets \vec{x})$ and
      $\omega_\tau(\vec{X}' \gets \vec{x}')$ or $Y_j$ is assigned a value in one
      of $\omega_\tau(\vec{X} \gets \vec{x})$ and
      $\omega_\tau(\vec{X}' \gets \vec{x}')$ and not in the other.   
      %
      By definition, this means that for all variables $W \in \vec{Z}_j$, we have that $W \prec_{\vec{u}_L} V$. Furthermore, at
      least one of these variables $W$ either gets different values in
      $\vec{X} \gets 
      \vec{x}$ and 
      $\vec{X}' \gets \vec{x}'$, or is assigned a value in one
      of $\vec{X} \gets \vec{x}$ and
      $\vec{X}' \gets \vec{x}'$ and not in the other.   

      This concludes the proof.
      \eprf

      \commentout{
        \commentout{
          {\bf Composite Interventions:}
          One of the challenges for multi-level causal models is to ensure that 
          there are no counterintuitive results for composite interventions,
          i.e.\ for interventions that intervene on more than one variable. The
          issues that can arise are nicely illustrated in examples 7 and 8 in
          Rubinstein et al. \citeyear{Rub17}. They solve this concern by
          demanding that $\omega$ be \emph{order-preserving}, i.e.\ that the
          partial order over intervention sets is preserved in both the low and
          high level models. In the framework here, order-preservingness is
          guaranteed as well, since it can be shown that $\omega_{\tau}$
          (Definition 2.3) is order-preserving. Specifically,
          order-preservingness of interventions follows from how $\omega_{\tau}$
          is defined in terms of restrictions $Rst(.)$: 

          Say, $\omega_{\tau}(X \gets x)=Y \gets y$. Now, say we have $X \gets x   \preceq   X' \gets x'$, where ``$\preceq$'' is the natural order as discussed by  Rubinstein et al. \citeyear{Rub17}. Say, $\omega_{\tau}(X' \gets x')=Y' \gets y'$.  We need to prove that $Y \gets y  \preceq  Y' \gets y'$ to satisfy order-preservingness.

          By definition of $Rst$,  $Rst(x') \subseteq Rst(x)$. So $\tau(Rst(x')) \subseteq \tau(Rst(x))$. But we know that $\tau(Rst(x))=Rst(y)$ and $\tau(Rst(x')=Rst(y')$, and therefore $Rst(y') \subseteq Rst(y)$. But that means precisely that $Y \gets y  \preceq  Y' \gets y'$.

          One way to think about the present approach is that abstractions already place stronger constraints on the relation between the low-level and high-level models than the transformations in  Rubinstein et al. \citeyear{Rub17}. Since we in general think that abstractions are more appropriate for causal models, we consider this to be an advantage of the present framework. 
        }%
        {\bf Composite Interventions and $\omega$:}
        One of the challenges for multi-level causal models is to ensure that
        there are no counterintuitive results for composite interventions,
        that is, for interventions that intervene on more than one variable. The
        issues that can arise are nicely illustrated by Rubstein et al's
        citeyear{Rub17} Examples 7 and 8. in  They deal with this concern by
        requiring that $\omega$ be \emph{order-preserving}, that is, that the
        partial order over intervention sets is preserved in both the low and
        high level models. This aspect is a key component of their notion of
        an exact $\tau$-transformation. 

        Beckers and Halpern~\citeyear{BH19} introduced the notion of
        $\tau$-abstraction as a more restricted form of an 
        exact $\tau$-transformation.  

        Specifically, order-preservingness of interventions follows from how
        $\omega_{\tau}$ is defined in terms of restrictions $Rst(.)$: 

        Say, $\omega_{\tau}(X \gets x)=Y \gets y$. Now, say we have $X \gets x
        \preceq   X' \gets x'$, where ``$\preceq$'' is the natural order as
        discussed by  Rubinstein et al. \citeyear{Rub17}. Say,
        $\omega_{\tau}(X' \gets x')=Y' \gets y'$.  We need to prove that $Y
        \gets y  \preceq  Y' \gets y'$ to satisfy order-preservingness. 

        By definition of $Rst$,  $Rst(x') \subseteq Rst(x)$. So $\tau(Rst(x')) \subseteq \tau(Rst(x))$. But we know that $\tau(Rst(x))=Rst(y)$ and $\tau(Rst(x')=Rst(y')$, and therefore $Rst(y') \subseteq Rst(y)$. But that means precisely that $Y \gets y  \preceq  Y' \gets y'$.
      }
    }

   \subsubsection*{Acknowledgements}
    Beckers was supported by 
     the
     grant ERC-2013- CoG project REINS
     616512.
    Eberhardt was supported in part by NSF grants 1564330 and BCS-1845958, and HRGC grant 13601017.
    Halpern was supported in part by NSF 
    grants IIS-1703846 and IIS-1718108, ARO grant W911NF-17-1-0592, and a
    grant from the Open Philanthropy project. 
    We thank the 
    UAI reviewers for many useful comments.

    \bibliographystyle{chicago}
    \bibliography{joe}

\end{document}